\documentclass[review,number,sort&compress]{elsarticle}
\usepackage{lineno}
\usepackage{amssymb}
\usepackage{color}
\usepackage{soul}
\usepackage{physics}% added to include \hat
\usepackage{subcaption}% added to include subplot
\usepackage{verbatim} % comments
\usepackage{graphicx} 
\usepackage[ansinew]{inputenc}
\usepackage{appendix}
\usepackage{algorithm}
\usepackage{listings}%grammar
\usepackage[noend]{algpseudocode}
\journal{Journal of Computers in Biology and Medicine}

\begin{document}

\begin{frontmatter}

% Homework 21/2/2023
% - David: SINDy chapter text, expressions 
% - Daniel: include latest versions of plots. Black dots. Slightly larger.
% - David, Daniel: Test for the results / discussion section
% - Inaki: Complete Intro
% - Manuel: Finish State-of-the-art / Related Work
% - Gabriel: Editing once we have all the content complete.
%\shorttitle{Learning Difference Equations}

% Short author
%\shortauthors{Parra et~al.}

\title{Learning Difference Equations with Structured Grammatical Evolution for  Postprandial Glycaemia Prediction}
 % \title{Comparing different techniques for Predicting Postprandial Glycaemia}

\author[addr1]{Daniel Parra }
\author[addr2]{David Joedicke }
\author[addr1]{J. Manuel Velasco}
\author[addr2]{Gabriel Kronberger}
\author[addr1]{J. Ignacio Hidalgo}

\address[addr1]{Department of Computer Architecture and Automatics, Universidad Complutense de Madrid Madrid, Spain. \{dparra02, mvelascc, hidalgo\}@ucm.es}
\address[addr2]{University of Applied Sciences of Upper Austria, Hagenberg, Austria. \{David.Joedicke, Gabriel.Kronberger\}@fh-hagenberg.at}

%\address[addr2]{}
%\address[addr3]{}

\begin{abstract}

%%%%%%%%%%%%%%%%%%%%%%%%%%%%%%%%

People with diabetes must carefully monitor their blood glucose levels, especially after eating. Blood glucose regulation requires a proper combination of food intake and insulin boluses. Glucose prediction is vital to avoid dangerous post-meal complications in treating individuals with diabetes. Although traditional methods, such as artificial neural networks, have shown high accuracy rates, sometimes they are not suitable for developing personalised treatments by physicians due to their lack of interpretability.
In this study, we propose a novel glucose prediction method emphasising interpretability: Interpretable Sparse Identification by Grammatical Evolution. Combined with a previous clustering stage, our approach provides finite difference equations to predict postprandial glucose levels up to two hours after meals. We divide the dataset into four-hour segments and perform clustering based on blood glucose values for the two-hour window before the meal. Prediction models are trained for each cluster for the two-hour  windows after meals, allowing predictions in 15-minute steps, yielding up to eight predictions at different time horizons.
Prediction safety was evaluated based on Parkes Error Grid regions. Our technique produces safe predictions through explainable expressions, avoiding zones D (0.2\% average) and E (0\%) and reducing predictions on zone C (6.2\%). In addition, our proposal has slightly better accuracy than other techniques, including sparse identification of non-linear dynamics and artificial neural networks. The results demonstrate that our proposal provides interpretable solutions without sacrificing prediction accuracy, offering a promising approach to glucose prediction in diabetes management that balances accuracy, interpretability, and computational efficiency.
%%%%%%%%%%%%%%%%%%%%%%%%%%%%%%%%

\end{abstract}

\begin{figure}[ht!]
    \centering
    \includegraphics[width=\textwidth]{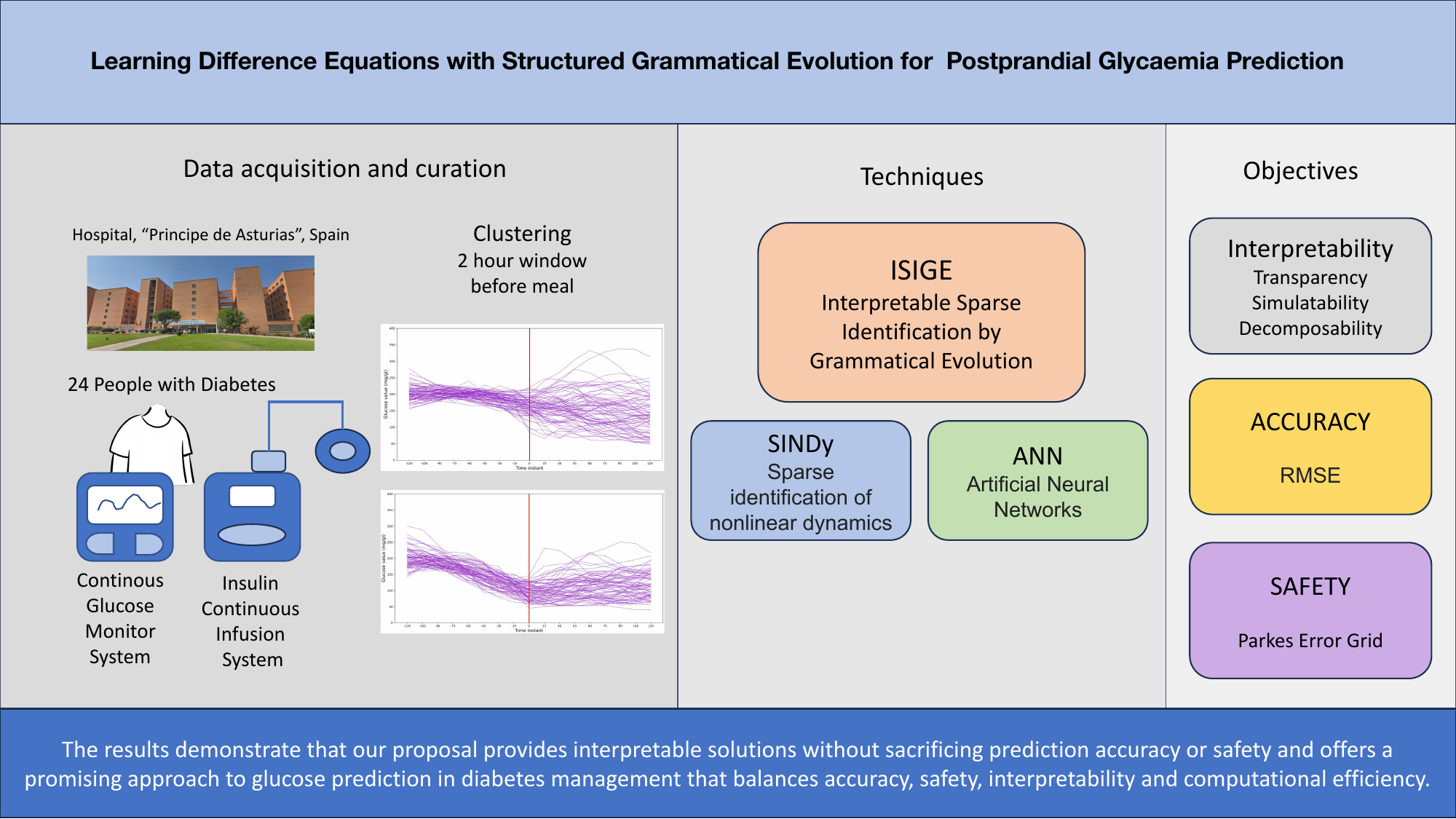}
    \caption{Graphical Abstract}
    %\caption{Work flow diagram of our methodology. Data from 24 participants is split into segments spanning four hours (2h before and 2h after the meal). The segments are clustered into 15 clusters based on the blood glucose values 2h before the meal. Predictions are made for a forecast window of 2h after the meal using difference equations generated with ISIGE, SINDy and three different NN models.}
    \label{fig:Graphical_Abstract}

\end{figure}%\marginpar{Use correct capitalization in figure.}-----------------------------------------------------------------------------------------------

\begin{highlights}
\item This paper proposes a novel approach to glucose prediction in diabetes management that emphasises interpretability: Interpretable Sparse Identification by Grammatical Evolution (ISIGE)
\item The proposed technique combines clustering with ISIGE to obtain finite difference equations that predict postprandial glucose levels up to two hours after meals.
\item We have employed data from 24 different participants with diabetes mellitus type-I that was clustered based on blood glucose values for the two-hour window before the meal
\item Although ANN is one of the best-performing techniques for glucose prediction and one of the most commonly used methods in this field, in our study, both SINDy and ISIGE obtained better results in most clusters.
\end{highlights}

\end{frontmatter}

\section{Introduction}
\label{cap:Introduction}

More than 450 million people suffer from diabetes. There are two main types of diabetes: type I is an autoimmune disease that causes the
destruction of insulin-producing cells (beta cells) of the pancreas, while type II appears when there is resistance to insulin action. 

Insulin-dependent diabetes patients need to estimate, or even better to predict their blood glucose levels in the near term to manage their condition and prevent complications. Predicting glucose levels can help individuals make informed decisions about their diet, exercise, and especially the insulin and medication they use to maintain their blood glucose within a healthy range. The last decade has seen the rapid spread of new and reliable continuous glucose monitoring systems (CGM), which provide real-time glucose readings and trend data to help individuals adjust
their treatment plan accordingly. 

The availability of data from CGMs has lead the research in the field. Great efforts have been made in the search for accurate glucose prediction models. Some of them are black-box models~\cite{mirshekarian2019lstms}, others are based on analytical models~\cite{Cobelli1983-gq}, and most of them provide predictions for a time horizon from 15 to 120 minutes~\cite{FELIZARDO2021glucose_prediction}. Among them, symbolic regression (SR) \cite{CONTADOR2021107609} techniques and artificial neural networks (ANNs) obtained very good performance \cite{Tena2021BGpredNN}.

ANNs have become a popular tool for glucose prediction in the management of diabetes due to their ability to capture complex, non-linear relationships between input variables and glucose levels. However, one of the key challenges in using ANNs for glucose prediction is the lack of interpretability of the solutions they provide. ANNs can be viewed as black-box models, making it difficult to understand how they arrive at their predictions and limiting their usefulness in clinical decision making. This problem of interpretability has led researchers to explore alternative explainable AI techniques to provide both accurate forecasts and insight for clinicians. 

This work aims to explore techniques for deriving finite difference equations that accurately represent the dynamics of blood glucose levels, with the benefit of obtaining interpretable models that can serve to aid the work not only of people with diabetes, but also of diabetes clinicians. To achieve this interpretability, we have developed a variant of grammatical evolution (GE) that produces sparse solutions: Interpretable Sparse Identification by Grammatical Evolution (ISIGE). This technique seeks to integrate the good results in the last years of Grammatical Evolution for blood glucose prediction,\cite{Hidalgo2018} and the advantages of a recently proposed method: sparse identification of non-linear dynamics (SINDy).

SINDy is an optimization-based method that constructs a sparse model of the dynamics of a system as a linear model with a set of non-linear base functions. To achieve sparsity, SINDy adds a sparsity-promoting regularization term which penalizes models with too many non-zero coefficients, forcing the algorithm to select a smaller subset of variables and interactions most relevant to the dynamics of the system. This helps to avoid overfitting, improve the interpretability of the model, and reduce computational complexity \cite{brunton_kutz_2022}. Conceptually, the approach is similar to fast function extraction (FFX)~\cite{McConaghy2011} but focuses on system dynamics.

In addition, to refine the models and predictions, we employ a clustering approach to group the glucose time series before meals which give us several scenarios for prediction. Due to the complexity of this situation and the need for a technique with good performance characteristics, we have implemented ISIGE using Dynamic Structured GE (DSGE). In addition, to reduce the risk of premature convergence and to improve the diversity of the generated solutions, we have applied the $\epsilon$-\emph{lexicase} selection technique \cite{laCava2019eLexicase}. 

In this paper, we describe our technique and analyze the experimental results against SINDy and ANNs, in terms of precision, suitability for diabetes care and interpretability.

The  results of this study are significant in several ways:
\begin{itemize}
    \item The results demonstrate that the proposed ISIGE approach, together with SINDy, outperforms well tested ANNs in terms of prediction accuracy.
    \item ISIGE is the technique that provides the most accurate predictions. This is particularly important for glucose prediction in diabetes treatment, where accurate and safe predictions are crucial for clinical decision-making.
    \item The fact that ISIGE and SINDy show similar interpretability suggests that ISIGE can provide interpretable solutions without sacrificing prediction performance and safety.
\end{itemize}

These results offer a promising new approach to glucose prediction in diabetes management that balances accuracy, interpretability and computational efficiency.

%The rest of the paper is organized as follows. Section \ref{sec:GP_relatedwork} reviews the literature on hypoglycemic prediction with machine learning (ML) techniques. Section \ref{sec:ISIGE} explains our approach (), \ref{sec:Application_Description} , the workflow, and the main techniques used in this research, whereas Section \ref{sec:results} describes the experimental setup, the data used, and the experimental results obtained with our proposal. Conclusions are given in Section \ref{sec:conclusion}.

The subsequent sections of this article are structured as follows: Section \ref{sec:GP_relatedwork} presents an overview of the existing literature on utilizing machine learning (ML) techniques for hypoglycemia prediction. In Section \ref{sec:ISIGE}, our approach is elucidated, encompassing the specific features of the grammar used in our version of DSGE, together with the iterative numerical evaluation algorithm used for the fitness computation. Section \ref{sec:Application_Description} outlines the workflow and primary techniques employed in this research as a point of reference. The experimental setup, data utilized, and the resulting outcomes obtained from our proposed approach are detailed in Section \ref{sec:results}. Finally, Section \ref{sec:conclusion} provides the concluding remarks of this study.

\section{Related work}	\label{sec:GP_relatedwork}

Dynamic models for glucose prediction are mathematical models that can simulate and predict behavior of blood glucose levels over time. They have been used in a variety of applications, such as personalized glucose control algorithms, prediction of hypoglycemia events, and optimization of insulin therapy.

There are three types of dynamic models for glucose prediction:

\begin{itemize}
    \item Physiological models \cite{Cobelli1983-gq, Dalla_Man2007-au, Dalla_Man2006-ee}: These models use a set of differential equations to simulate the dynamics of glucose and insulin in the body. Physiological models are based on our current understanding of the complex interactions between glucose and insulin in the body.
    
    \item Data-driven models \cite{Felizardo2021, Zhu2018ADL, li2019convolutional, lobo2022, Hamdi2017, Zecchin2012}: These are mathematical models that are constructed based on data obtained from individuals with diabetes. These models use statistical and machine learning techniques, to analyze the data and identify patterns that can be used to predict future glucose levels. Different models have been developed in conjunction with CGM systems to make short-term or long-term predictions.
    
    \item Hybrid models \cite{Hovorka2004, Contreras2017, Liu2019-az}: These models combine physiological and data-driven models to take advantage of the strengths of both types of models to improve the accuracy of glucose predictions. 
\end{itemize}

As we can see, blood glucose prediction is a very active field of research, and although significant progress has been made, several challenges remain to be overcome, including:
\begin{itemize}
    \item Individual differences: the metabolism of each person is unique, and blood glucose dynamics can vary significantly from individual to individual and from the situation of each person (e.g., changes during pregnancy, illness requiring medication administration, etc.). 
    \item Delayed response: There is always a variable delayed response between insulin administration or carbohydrate intake and blood glucose value. 
    \item Sensor limitations: The accuracy and reliability of glucose sensors can vary over their lifetime, so that the resulting noise can affect the data quality used for training and model prediction.
    \item Complex dynamics: Blood glucose dynamics are influenced by many factors that can interact in complex ways, such as food intake, physical activity, medication, sleep, and stress.
\end{itemize}

This paper addresses these challenges using two techniques that we consider particularly suitable for glucose prediction:  structured grammatical evolution (SGE) and SINDy. These two approaches can handle large and complex datasets with noisy and incomplete data, automatically extract the most relevant inputs, and identify the main mathematical functions to describe glucose dynamics. Additionally, they generate interpretable models that provide the opportunity to be analyzed, allowing researchers better to understand the underlying dynamics of blood glucose regulation.

SINDy \cite{brunton2016discovering, brunton_kutz_2022} is a data-driven approach used to discover governing equations that describe the dynamics of a system. Although it is a relatively new approach several research papers have been published addressing problems on different fields including physics \cite{Schaeffer2017-bu}, network biology \cite{mangan2016} or chemical kinetics \cite{KuntzWilson2022}.

The primary reference for GE applied to
glucose forecasting is the book chapter from Hidalgo et
al. \cite{Hidalgo2018}, where a new approach for
identifying mathematical models that can predict blood glucose levels
in people with diabetes using GE is proposed. The approach was evaluated using data from
the OhioT1DM dataset \cite{Marling2020-nu} containing glucose and
physiological data from 43 people with diabetes. The results showed that
the proposed system outperformed several baseline approaches regarding
accuracy and interpretability, including linear regression, ARIMA
models, and ANNs.

The lead article in the field of SGE is~\cite{nuno2019}.
This paper presents a novel
approach to predict glucose levels in people with diabetes using
SGE, a variant of GE that encodes the chromosomes of individuals using a list of
values corresponding to the productions of each non-terminal symbol to
increase the locality of genetic operators. Different selection schemes can be used for SGE.

Spector \cite{Spector2012} presented a preliminary report
with a novel idea: lexicase selection. Lexicase selection has been applied to many optimisation problems. 
Helmuth et al. \cite{helmuth2015, helmuth2016lexicase} used lexicase
selection to evolve models for several problems, being one of them
predicting blood pressure based on a set of physiological features.
LaCava and Spector (2016) \cite{cava2016}, propose a variant (called
Epsilon-lexicase) focused on continuous fitness landscapes that
produces better results for symbolic regression problems. Spector et
al. (2018) \cite{spector2018predicting} used lexicase selection to
evolve models for predicting protein-ligand binding affinity. The work
from Stanton et al. \cite{stanton2022} used lexicase selection to
evolve robot controllers for tasks such as navigation and object
manipulation.

\section{Methods and techniques} 
\label{sec:ISIGE}
A finite difference equation (FDE) is a mathematical expression that describes the difference between a variable $y$ at two discrete time points \cite{sharkovsky2012difference}. A FDE takes the form of Equation \ref{eq:ed_example}, where $\vec{x}$ represents a set of input variables involved in the equation, $y$ represents the target variable, and $\theta$ are optional calibration parameters. 
\begin{equation}
\label{eq:ed_example}
    \Delta y(t)=  y(t+\Delta{t})-y(t)=f(\vec{x}(t),y(t), \theta )
\end{equation}

FDEs are classified based on their properties, such as linearity, nonlinearity, and order, which is the highest difference in time steps that explicitly appear in the equation. For example, Equation \ref{eq:ed_example} is a first-order equation, while Equation \ref{eq:ed_example2} is a second-order equation as it involves the difference between $y(t+2)$ and $y(t)$.
\begin{equation}
\label{eq:ed_example2}
    y(t+2\Delta{t})-y(t)=f(\vec{x}(t))
\end{equation}

\begin{table}
\centering
  \begin{tabular}{c|l}
    Variable & Description \\
    \hline
    $B_I$ & Basal insulin \\
    $I_B$ & Insulin bolus. \\
    $F_{\text{ch}}$ & Carbohydrate intake.\\
    HR & Heart rate\\
    $G$ & Glucose concentration.\\
    $C$ & Calories burned\\
    $S$ & Steps.
  \end{tabular}
  \caption{Input variables in this study}
  \label{tab:variables}
\end{table}

Due to their simplicity, finite difference equations are handy for modelling dynamic processes which are measured with constant frequency (equidistant time steps), such as blood glucose levels measured by CGM. Equation \ref{eq:glucose_FDE}, express the dynamics of glucose values as a finite difference equation problem. The description of the input variables is given in Table \ref{tab:variables}.

\begin{equation}
    \Delta G(t) = G(t+\Delta_t) - G(t) = f(G(t), B_I(t), I_B(t), F_{ch}(t), HR(t), C(t), S(t))
    \label{eq:glucose_FDE}
\end{equation}

As is often the case in physical systems, we assume that in the glucose regulation system, only a few terms are relevant in defining its dynamics. Therefore, we can consider the governing equations are scattered in a high-dimensional non-linear function space. In this way, we aim to create a simplified model that accurately captures the essential dynamics of the system while minimising complexity. The SINDy algorithm is particularly suitable for this purpose. In subsection \ref{subsec:SINDy}, we briefly explain how SINDy works. Then, in subsection \ref{subsec:ISIGE}, we present our proposal, ISIGE, which seeks to incorporate the benefits of SINDy within the paradigm of GE. 

\color{black}
\subsection{Sparse Identification of Nonlinear Dynamics (SINDy)}
\label{subsec:SINDy}

\begin{algorithm}[h]
\caption{SINDy algorithm for Finite Difference Equations}\label{alg:SINDY_FD}
\begin{algorithmic}[1]
%\Procedure{SINDY}{$\mathbf{X},\Delta\_t, \lambda$}
\Procedure{SINDy}{$\mathbf{X},\Delta(t),\lambda$}
\State $\mathbf{X}, \mathbf{\dot{X}}  \gets$ History States Variables %$\mathbf{\Theta}(\mathbf{X})$
\State $\mathbf{\Theta}(\mathbf{X}) \gets$ candidate nonlinear functions library matrix %$\mathbf{\Theta}(\mathbf{X})$
\State $\mathbf{\Theta}(\mathbf{x}^T) \gets $ vector of symbolic functions
\State $\mathbf{x}^{P_2} \gets$ quadratic non-linearities in the state x
\State $\mathbf{\Theta}\lambda \gets$ add regularization term to $\mathbf{\Theta}$ (LASSO)
\color{black}
%\State $\mathbf{\dot{X}} \gets$ compute finite difference approximation of $\frac{d\mathbf{X}}{dt}$
\State $\boldsymbol{\Xi} \gets \boldsymbol{\xi}_1, \boldsymbol{\xi}_2, \boldsymbol{\xi}_3, ...$  sparse vectors of coefficients
\State $\boldsymbol{\xi}k \gets$ $\mathbf{\dot{X}} = \mathbf{\Theta}(\mathbf{X}) \boldsymbol{\Xi}$, solve sparse regression
\State Return $\mathbf{\dot{x}}_k = \mathbf{f}_k(\mathbf{x}) = \mathbf{\Theta}(\mathbf{x}^{P_2})\boldsymbol{\xi}_k$
\EndProcedure
\end{algorithmic}
\end{algorithm}

SINDy is a data-driven method for discovering the governing equations of a dynamical system from time series data  \cite{brunton2016discovering, brunton_kutz_2022}. In Algorithm \ref{alg:SINDY_FD}, we have summarized the main steps for SINDy.

\begin{itemize}
    \item Data gathering (line 2): The first step is to collect data from the system that we want to model. The data are time-series measurements of the state variables of the system (Table \ref{tab:variables}).
    \item Construct a library of candidate functions (line 3): The next step is to construct a library of candidate functions that could potentially be part of the governing equations. In the general SINDy algorithm, these functions could include polynomials, trigonometric functions, exponentials, etc, depending on the system being modeled. In this work and based in our prior knowledge of the system, we have chosen up to $X^{P_2}$, quadratic non-linearities of $X$ (Equation \ref{eq:matrix_XP2}).

\begin{equation}
\textbf{$\Theta$}(X) = \left[
\begin{array}{cccccccc}
\vdots& \vdots&\vdots&\vdots&\\
 1 & X & \vdots & X^{P_2} \\
\vdots& \vdots&\vdots&\vdots&\\
\end{array}
\label{eq:matrix_XP2}
\right]
\end{equation}

    \item The development of all the linear combinations (line 5), are grouped in Equation \ref{eq:XP2}, where $t_0$ represents the first time step, $t_1$ the subsequent time step, and so on.

\begin{equation}
\label{eq:XP2}
\textbf{$X^{P_2}$} = \left[
\begin{array}{ccccc}
G(t_0)\cdot B_I(t_0)& G(t_0)\cdot I_B(t_0)& G(t_0)\cdot F_CH(t_0) &  \cdot\cdot\cdot & F_CH^2(t_0)\\
G(t_1)\cdot B_I(t_1)& G(t_1)\cdot I_B(t_1)& G(t_1)\cdot F_CH(t_1) &  \cdot\cdot\cdot & F_CH^2(t_1)\\
\vdots & \vdots & \vdots &\vdots & \vdots  \\
G(t_n)\cdot B_I(t_n)& G(t_n)\cdot I_B(t_n)& G(t_n)\cdot F_CH(t_n) &  \cdot\cdot\cdot & F_CH^2(t_n)\\
\end{array}
\right]
\end{equation}

    \item Add regularization (line 6): SINDy uses sequential threshold ridge regression as a regularization. The objective function  $||\Theta \xi - \dot{X}_t||^2_2 + \lambda ||\xi||^2_2$ is minimized by iteratively performing least squares and masking out elements of the weight array $\xi$ that are below a given threshold $\lambda$. In this paper we used $\lambda = 0.5$ which tests have shown to produce the best results.
    
    \item Solving the optimization problem (lines 7-8): through a sparse regression algorithm, we identify the combination of terms that best describes the observed dynamics of the system.
    \item Once the sparse coefficients have been found (line 9), we have the governing equations of the system.
\end{itemize}

\begin{comment}
Starting from a basic dynamical system (Equation \ref{eq:1}), we
consider that future glucose values are based primarily on the current
glucose value, the basal rate of insulin administration, carbohydrate
intake and bolus insulin injections.
At this point, we use a library of candidate functions to create a set of
possible terms (Equation
\ref{eq:3}) in the governing equations. Then, through a sparse
regression algorithm, we identify the combination of terms
that best describes the observed dynamics of the system.

{\color{red} GKR: equations are incorrect as we only predict blood glucose and use measured values for the other variables as inputs
\begin{eqnarray}
  \dv{}{t} \vec{x}(t) =\vec{f}(\vec{x}(t))& \label{eq:1} \\
  \vec{x}(t) =[G(t),B_I(t),I_B(t),F_{ch}(t)],&  \vec{x}(t) \in R^n \label{eq:2}
\end{eqnarray}
}

%{\color{red}GKR: Equation 5 for the full feature matrix not relevant here as we are not using trigonometic or P4 terms

Where $X^{P_2}$ represents the quadratic non-linearities of $X$,
$X^{P_3}$ the cubic non-linearities and so on (in this work we only
use up to $X^{P_2}$).
%}

{\color{red}GKR: this equation is not really useful as we do not know the definition of G, Bi, IB, FcH.
After application to our problem we obtain:

}

Where $t_0$ represents the first time step, $t_1$ the subsequent time step, and so on.
\end{comment}

%%%%%%%%%%%%%%%%%%%%%%%%%%%%%%%%%%%%%%%%%%%%%%%

\subsection{Interpretable Sparse Identification by Grammatical Evolution} 
\label{subsec:ISIGE}

This study introduces a novel approach, ISIGE, to model blood glucose dynamics using FDEs. Specifically, we utilize DSGE, a variant of GE, to obtain $\hat{G}(t+\Delta (t))$ as expressed in equation \ref{eq:Gt+1} which comes from the expression \ref{eq:glucose_FDE}.

\begin{equation}
\hat{G}(t+\Delta (t)) =  G(t) + \Delta G(t) = G(t) + f(\Vec{x}(t))
\label{eq:Gt+1}
\end{equation}

 One of the benefits of using DSGE is the ability to obtain explainable expressions, which enables us to analyze the impact and significance of different components in the system. Therefore, our proposed approach is expected to provide interpretable and efficient models for blood glucose dynamics.
 DSGE addresses two limitations of GE. First, it overcomes the low locality problem by ensuring that a slight change in the genotype results in a corresponding small change in the phenotype, thus ensuring high locality. Second, it eliminates redundancy, which occurs when different genotypes produce the same phenotype, by creating a one-to-one mapping between the genotype and the non-terminal.
To achieve this, DSGE employs variable-size lists instead of the fixed-size lists used in GE.

\subsubsection{$\epsilon$-Lexicase selection}
\label{subsubsec:lexicase}

The characteristics of our data are explained in depth in Section \ref{subsec:clustering}. Broadly speaking, we have grouped the data into clusters comprising multiple 4-hour glucose time series segments. The behaviour of these segments is highly varied (Figure \ref{fig:clusters}), and if the selection mechanism of the evolutionary algorithm fails to take this into account, many individuals who do not achieve high fitness in all cases will be rejected. Because of this, we think of lexicase (and, more specifically, $\epsilon$-Lexicase) as a selection mechanism. 

Lexicase selection \cite{helmuth2014lexicase} is a powerful search strategy used in evolutionary computation to overcome the problem of premature convergence. It works by selecting individuals that perform well in a randomly chosen subset of the fitness cases and repeating the process with the remaining fitness cases until only one individual is left. However, this method may become too selective, leading to limited search space exploration and the loss of diversity. To address this issue, $\epsilon$-Lexicase \cite{laCava2019eLexicase} selection was proposed, introducing a tolerance parameter, $\epsilon$. This parameter selects individuals that perform well within a certain threshold of the best-performing individuals in each subset. This approach allows for a more comprehensive search space exploration while preserving good individuals.

Algorithm \ref{algo:eLDSGE} outlines the process of obtaining difference equations using  ISIGE. We load the datasets and problem properties, then create an initial population and start the evolutionary process. We obtain the phenotypes (expressions) of each generation by decoding the population with the grammar. Using $\epsilon$-Lexicase selection, we determine the parents for the next population. We generate a new population by crossing and mutating the selected parents.

\begin{algorithm}[]
\caption{ ISIGE}
\label{algo:eLDSGE}
\begin{algorithmic}[1]

\Procedure{ ISIGE}{$grammar,datasets,properties$}       
    \State Load properties
    \State Load Datasets
    \State pop = Generate population(len\_pop) 
    \For{$i \gets 0$ to ($N\_generations$ - 1)}  
        \For{$j \gets 0$ to ($len\_pop$ - 1)}
            \State $\hat{G}(t+\Delta{t})_j$ = decode(grammar, $pop_j$)
        \EndFor
        
        \State pop= $\epsilon$-Lexicase selection (datasets, pop , $f_j(\Vec{x}(t)))$
        \State pop = Crossover(pop)
        \State pop = Mutation(pop)
    \EndFor
\EndProcedure

\end{algorithmic}
\end{algorithm}
Where $\hat{G}(t+\Delta{t})_j$ comes from Equation \ref{eq:Gt+1}
\subsubsection{Iterative Numerical Evaluation}
\label{sec:INE}

During the $\epsilon$-Lexicase selection process, we evaluate individuals from the current subpopulation using a specific dataset for each iteration. 
One difference in our approach is the application of an iterative numerical method to solve Equation \ref{eq:Gt+1}. This equation calculates the glucose concentration for the next time step ($\hat{G}(t + \Delta t)$). We have developed algorithm \ref{algo:INE}, named Iterative Numerical Evaluation (INE), to implement this iterative numerical method. In our study, each time step ($t_i$) represents samples taken every 15 minutes within two hours after a meal. We evaluate the predicted glucose value at the next time step ($\hat{G}(t_{i+1})$) using various variables of the system ($\Vec{x_k}(t_i)$.) from the data set, in the case of glucose, only the first value is taken, and for the remaining iterations, the predicted value obtained in the previous step is used. Once predictions are made for all time instances, we calculate the Root Mean Square Error (RMSE) by comparing the predicted glucose values with the actual values using Equation \ref{eq:rmse}. The resulting RMSE serves as the fitness measure for that individual on the specific dataset, indicating its performance for that dataset.

\begin{equation}
\label{eq:rmse}
\text{RMSE} = \sqrt{\frac{1}{N}\sum_{t=1}^N\left(\hat{G}_{t}-G_{t}\right)^2}
\end{equation}

In Equation \ref{eq:rmse}, $N$ represents the total number of time instances, $\hat{G}_t$ denotes the predicted glucose value at time $t$, and $G_t$ represents the actual glucose value at time $t$.

 The iterative numerical evaluation, algorithm \ref{algo:INE}, takes as input a function $f_j$ and a dataset $D_k$. It proceeds as follows:

\begin{enumerate}
    \item Initialize the predicted glucose value at the starting time ($t_0$) as the actual glucose value from the dataset ($G_k(t_0)$).
    \item Iterate through each time step ($t_i$) from 0 to the total number of points in the dataset minus one.
    \item Update the predicted glucose value at the next time step ($t_{i+1}$) by adding the function $f_j$ applied to the variables $\Vec{x_k}(t_i)$.
    \item Calculate the fitness $F_j$ of individual $j$ for the dataset $k$ by comparing the predicted glucose values ($\hat{G}$) with the actual glucose values ($G_k$) using the Root Mean Square Error (RMSE) measure, equation \ref{eq:rmse}.
    \item Return the fitness $F_j$.
\end{enumerate}

\begin{algorithm}[]
\caption{Iterative numerical evaluation}
\label{algo:INE}
\begin{algorithmic}[1]
\Procedure{Iterative\_numerical\_evaluation}{$f_j$,$D_k$}       
    \State $\hat{G}(t_0)=G_k(t_0) $ 
    \For{$i \gets 0 $ to ($num\_points$ - 1)}
        \State $\hat{G}(t_{i+1})$ = $\hat{G}(t_i)+f_j (\Vec{x_k}(t_i))$
    \EndFor
        \State $F_j$=RMSE($\hat{G}$, $G_k$)
\State return $F_j$
\EndProcedure
\end{algorithmic}
\end{algorithm}

Where $G_k$ is the set of actual glucose values of $D_k$, $\hat{G}$ is the set of predicted values for the same points of $D_k$, $\Vec{x_k}(t_i)$ the set of variables of $D_k$ for time i and $F_j$ is the fitness of individual j for $D_k$.
%In our study, each time step ($t_i$) represents samples taken every 15 minutes within two hours after a meal. We evaluate the predicted glucose value at the next time step ($\hat{G}(t_{i+1})$) using various variables of the system ($\Vec{x_k}(t_i)$.), including glucose ($\hat{G}(t_i)$), basal insulin ($I_B(t_i)$), insulin bolus ($B_I(t_i)$), carbohydrate intake ($F_{ch}(t_i)$), heart rate ($HR(t_i)$), calories burned ($C(t_i)$), and step count ($S(t_i)$). These variables are obtained from the dataset $dataset_k$.
%In our study, each time step ($t_i$) represents samples taken every 15 minutes within two hours after a meal. We evaluate the predicted glucose value at the next time step ($\hat{G}(t_{i+1})$) using various variables of the system ($\Vec{x_k}(t_i)$.)   in subsequent iterations, the predicted value will be used.

%The INE algorithm (Algorithm \ref{algo:INE}) represents the iterative numerical evaluation process, where glucose values are predicted for each time step after a meal using a given function and dataset. The algorithm starts with the actual glucose value at the first time step and predicts subsequent values based on the previous prediction. 
This process is repeated for each individual in the subpopulation of this $\epsilon$-Lexicase selection iteration.

\subsubsection{Grammar}
\label{Grammar}
\begin{figure}
    \centering
    
    \includegraphics[width=\textwidth]{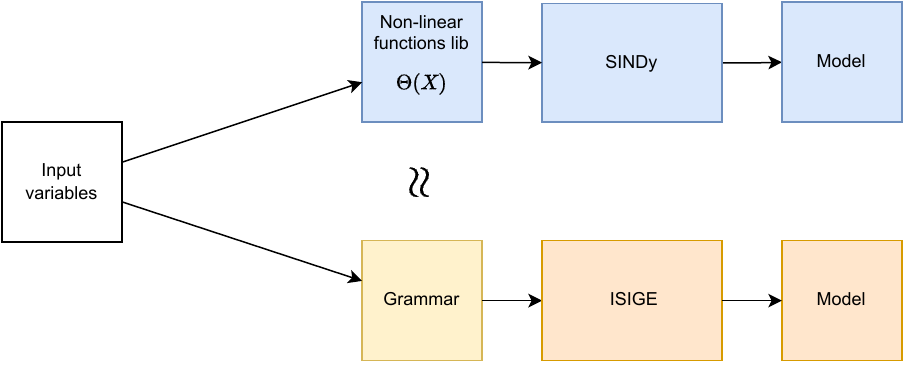}
    \caption{Common steps between SINDy and ISIGE, our proposal}
    \label{fig:SINDy_eLDSGE}
    %\vspace{-0.5cm}
\end{figure}

Creating a grammar to define the search space is essential in DSGE. This involves establishing how to generate constants, which operations to use, which structures to generate expressions and which variables to incorporate. In Figure \ref{fig:SINDy_eLDSGE}, we can see that the set of base functions used in the SINDy algorithm (subsection \ref{subsec:SINDy}) is equivalent to the ISIGE grammar, but there is an important consideration here. Due to the method used to evaluate the numerical expressions, it is only possible to calculate the values of the variables for one time step at each iteration.

%Figure \ref{fig:grammar} shows the grammar used for this work; among the most important elements, we find the initial definition of the expression $<$func$>$, in which it is established that all the expressions follow the form G + $<$ expr$>$, thus ensuring that we will obtain FDEs. Another element to highlight is the field about variables,$<$var$>$, where we find three parts, the individual variables, except G, which has already been made to appear, the first-order non-linearities, and the second-order non-linearities. In this way, getting closer to the ``library'' of functions used in the SINDy algorithm is possible. 

%\color{blue}
In Figure \ref{fig:grammar}, we present the grammar used in this work  in Backus Naur Form (BNF), which is crucial for obtaining the desired FDEs. One of the essential elements of this grammar is the initial definition of the $<$func$>$ expression, which establishes that all expressions must follow the form G + $<$ expr$>$, ensuring that the resulting expressions are in the FDE form. Another critical element is the $<$var$>$ field, which includes individual variables (excluding G, which has already been included), first-order non-linearities, and second-order non-linearities. This approach allows us to mimic the set of base functions used in the SINDy algorithm. Using this grammar, we can effectively generate diverse FDEs that describe the system behaviour and make accurate predictions.
%\color{black}

\begin{figure}[!htbp]
\begin{lstlisting}[basicstyle=\scriptsize\ttfamily,breaklines=true,frame=tb,mathescape]
# Model expression
<func> ::= G + <expr>
<expr> ::= (<expr> <op> <expr>) | (<cte> <op> <var>  <op> <expr>)
           | <var> | (-<var>) 
           | pow(<var>,<sign><exponent>)
           | (-pow(<var>,<sign><exponent>))
<var> ::= $B_I | I_B|F_{ch}| \text{HR} |C |S|$
          # First order 
          $G \cdot B_I|G \cdot I_B|G \cdot F_{ch}|G \cdot \text{HR}|G \cdot C|G \cdot S|$
          $B_I \cdot I_B|B_I \cdot F_{ch}|B_I \cdot \text{HR}|B_I \cdot C|B_I \cdot S|$          
          $I_B \cdot F_{ch}|I_B \cdot \text{HR}|I_B \cdot C|I_B \cdot S|$          
          $F_{ch} \cdot \text{HR}|F_{ch} \cdot C|F_{ch} \cdot S|$          
          $\text{HR} \cdot C|\text{HR} \cdot S|$          
          $C \cdot S|$
          # Second order
          $G \cdot G | B_I \cdot B_I | I_B \cdot I_B | F_{ch} \cdot F_{ch} | \text{HR} \cdot \text{HR} | C \cdot C | S \cdot S$          
<op> ::= +|-|$\cdot$ 
<cte> ::= <base> $\cdot$ pow(10,<sign><exponent>)
<base> ::= 1|2|3|4| $\cdot \cdot \cdot$ |99
<exponent> ::= 1|2|3|4|5|6|8|9
<sign> ::= +|-
\end{lstlisting}
\caption{Grammars used for ISIGE in BNF.}
    \label{fig:grammar}
\end{figure}

\section{Workflow} \label{sec:Application_Description}
\begin{figure}[ht!]
    \centering
    \includegraphics[width=\textwidth]{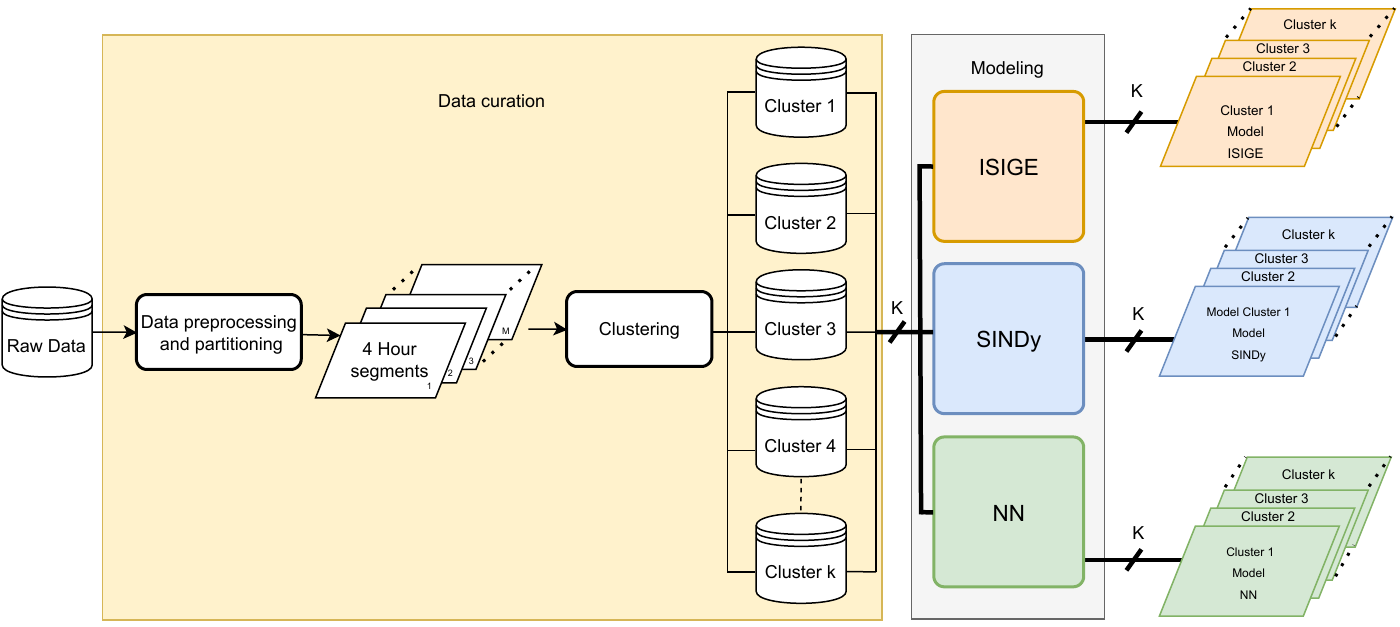}
    \caption{Workflow diagram of our methodology. The stages represented are: data acquisition, data preprocessing, data partitioning, clustering, modelling.}
    %\caption{Work flow diagram of our methodology. Data from 24 participants is split into segments spanning four hours (2h before and 2h after the meal). The segments are clustered into 15 clusters based on the blood glucose values 2h before the meal. Predictions are made for a forecast window of 2h after the meal using difference equations generated with ISIGE, SINDy and three different NN models.}
    \label{fig:methodology}

\end{figure}

Figure~\ref{fig:methodology} summarizes the workflow of the methodology applied in this work. It consists on several steps, further described in this section: data acquisition, data preprocessing, data partitioning, clustering, modelling, testing and results comparison.

%First, $M$ segments of four hours are generated from the raw data in a preprocessing step. The segments are assigned to clusters based on the glucose values in the first half of the segment. Once the clusters are obtained,each of the tested methods is used to produce a prediction model in the form of a difference equation for the second half of the segment.In the following, each of the steps is described in more detail.

  %\marginpar{Use correct capitalization in figure.}-----------------------------------------------------------------------------------------------

%Include one Figure with the description of the methodology.

\subsection{Data acquisition} \label{cap:raw_data}

For this study, we used data from 24 participants of the Hospital Universitario Pr\'{i}ncipe de Asturias, Madrid, Spain. Two different devices were used to obtain the raw data: A continuous glucose monitor (CGM) system (Free Style Libre sensor) and an activity monitoring wristband (Fitbit Ionic). The CGM measures interstitial glucose levels every 15 minutes, while the wristband records data on calories, steps and heart rate  at different time frequencies. 

In addition, information about insulin and carbohydrate intakes was recorded by two different methods depending on the insulin administration mode. Participants wearing an automatic insulin continuous infusion system (insulin pumps) obtain this information directly from the device (Medtronic or Roche systems). Participants under multiple doses of insulin  (MDI) therapy recorded  the information about basal insulin, insulin boluses, and carbohydrate intakes using a mobile application.

The studies involving human participants were reviewed and approved by Hospital Universitario Pr\'{i}ncipe de Asturias (Spain). The patients/participants provided their written informed consent to participate in this study.

%Among the participants, we found 11 men and 13 women aged between 18 and 61. The data can be divided into two groups: glucose data, obtained using a continuous glucose monitor (CGM) system, and exercise data, obtained using an activity monitoring device, specifically the Fitbit Ionic model. The data used in this study include basal insulin, insulin boluses, carbohydrate intake, calories and steps, among other variables.

%%%%%%%%%%%%%%%%%%%%%%%%%%%%%%%%%%%%%%%%%%%%%%%%
\subsection{Data Preprocessing} \label{cap:preprocessing}

% TODO: Data Basis and Recording --> I will prepare sth, Inaki checks

To use the recorded data for the modelling of the blood glucose
level several preprocessing steps had to be performed. This includes
both cleaning as well as feature engineering.

\subsubsection{Features for the absorption of carbohydrates and insulin}
\begin{figure}
\centering
  
  \begin{subfigure}{0.49\textwidth}
    \includegraphics[width=\textwidth]{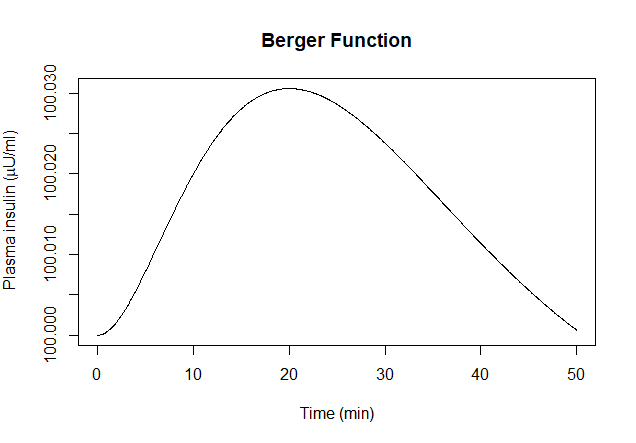}
    \caption{Possible plasma insulin evolution based on Berger equation}
    \label{fig:berger}
  \end{subfigure}
  \hfill
  \begin{subfigure}{0.49\textwidth}
    \includegraphics[width=\textwidth]{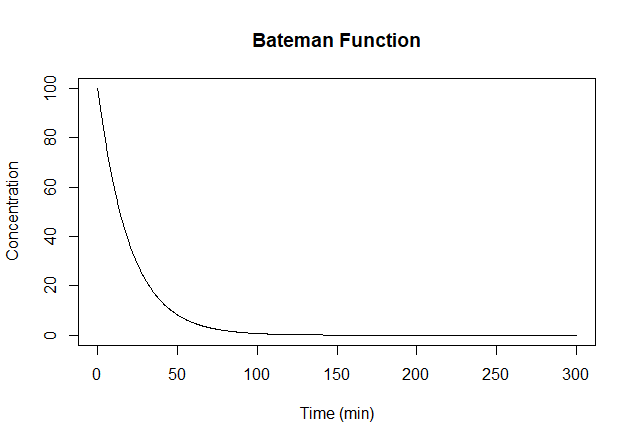}
    \caption{Possible insulin concentration evolution based on Bateman Function}
    \label{fig:bateman}
  \end{subfigure}
  \hfill
  \caption{Representation of Berger and Bateman function}
\label{fig:bergman_bateman}
\end{figure}

Since the dissolution of substances in the body occurs gradually,
we preprocessed both the reported insulin bolus and carbohydrate
values using two functions to spread the uptake over multiple observations: the Berger function
(Equation \ref{berger_fun} and Figure \ref{fig:berger}) \cite{berger1982absorption}  and the Bateman function (Equation \ref{bateman_fun} and Figure \ref{fig:bateman}) \cite{Garrett1994}.  \\

%%%%%%%%%%%%%%%%%%%
% manel

%%%%%%%%%%%%%%%%%%%

\newenvironment{conditions}
  {\par\vspace{\abovedisplayskip}\noindent\begin{tabular}{>{$}l<{$} @{${}={}$} l}}
  {\end{tabular}\par\vspace{\belowdisplayskip}}

The Berger function is
\begin{equation} \label{berger_fun}
   \frac{dA}{dt} = \frac{s \cdot t^s \cdot {(a \cdot D \cdot b)}^s \cdot D}{t \cdot ({(a \cdot D \cdot b)}^s+t^s)^2} - A
\end{equation}

where $A$ is the plasma insulin, $D$ the dose, $t$ the time after the consumption,  $s$ the mean absorption rate and $a$ and $b$ are parameters to characterize dependency on $D$. We use the parameter values $s=1.6$, $a=5.2$ and $b=41$ \cite{berger1982absorption}.

%where $A$ is the plasma insulin, $D$ the dose, $t$ the time after the consumption, {\color{red} $s$ the time course of absorption}, {\color{red} $s$ the mean absorption rate} %\marginpar{what is the time course of absorption? Reformulate} and $a$ and $b$ are parameters to characterize dependency on $D$. We use the parameter values $s=1.6$, $a=5.2$ and $b=41$ \cite{RefRequired}.

The Bateman function is
\begin{equation} \label{bateman_fun}
   c(t) = f \cdot \frac{D}{V} \cdot \frac{k_a}{k_a - k_e} \cdot (e^{-k_a \cdot t} - e^{-k_e \cdot t})
\end{equation}

where $k_a$ is the absorption rate, $k_e$ the elimination rate, $D$
the dose, $t$ the time since consumption, $V$ the volume of
distribution and $f$ the bioavailabilty. We used the parameter values
$k_a=0.1$, $k_e=0.2$, $V=0.5$ and
$f=0.5$ \cite{Garrett1994}.

Both functions convert an instant input such as the
insulin bolus as well as reported meals that the uptake of the
substance into the body does not take place abruptly but is distributed
over a longer period of time.

\subsubsection{Time-Shifts} \label{time-shifts}
We added time-shifted features for the  carbohydrates and the bolus values. The observations have a time shift of thirty minutes to the past, and the data points are recorded at fifteen-minute intervals.

%{\color{red} Observations are shifted for thirty minutes in the past, having a data-point every fifteen minutes.}\marginpar{Unclear, reformulate}

\subsubsection{Mean Values}
Since the heart rate and steps measurements every 15 minutes can change quickly, we smoothed them using a moving average with a window size of 30 minutes (two observations). 

\subsubsection{Creation of Segments}

We define a segment as the period of 2 hours before and after a reported meal (i.e. carbohydrate input). Segments may overlap if two meals are taken within a time span of two hours.
%{\color{red}
For each of those segments we check for the following constraints:

\begin{enumerate}
    \item Each of the parameter values for each time-step has to be greater than 0.
    \item We only allow interpolated data up to one hour.
    \item A change between time steps must not be more than 25\% of the previous value. 
\end{enumerate}
%}

\subsection{Clustering}
\label{subsec:clustering}
\begin{figure}
\centering
  
  \begin{subfigure}{0.49\textwidth}
    \includegraphics[width=\textwidth]{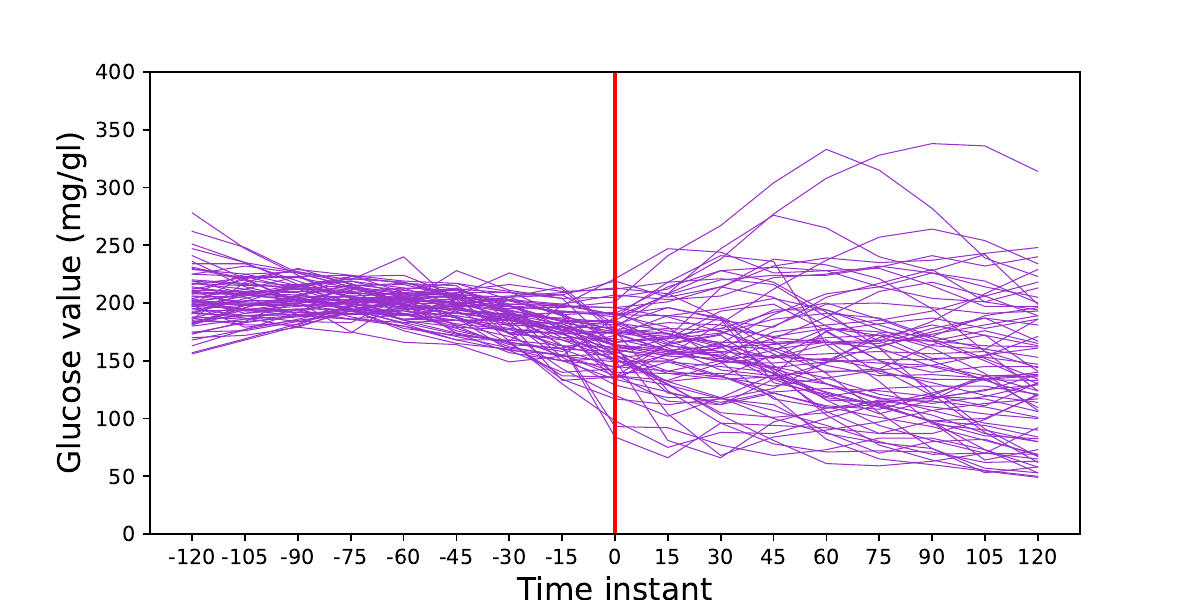}
    \caption{Cluster 4}
    \label{fig:cluster_4}
  \end{subfigure}
  \hfill
  \begin{subfigure}{0.49\textwidth}
    \includegraphics[width=\textwidth]{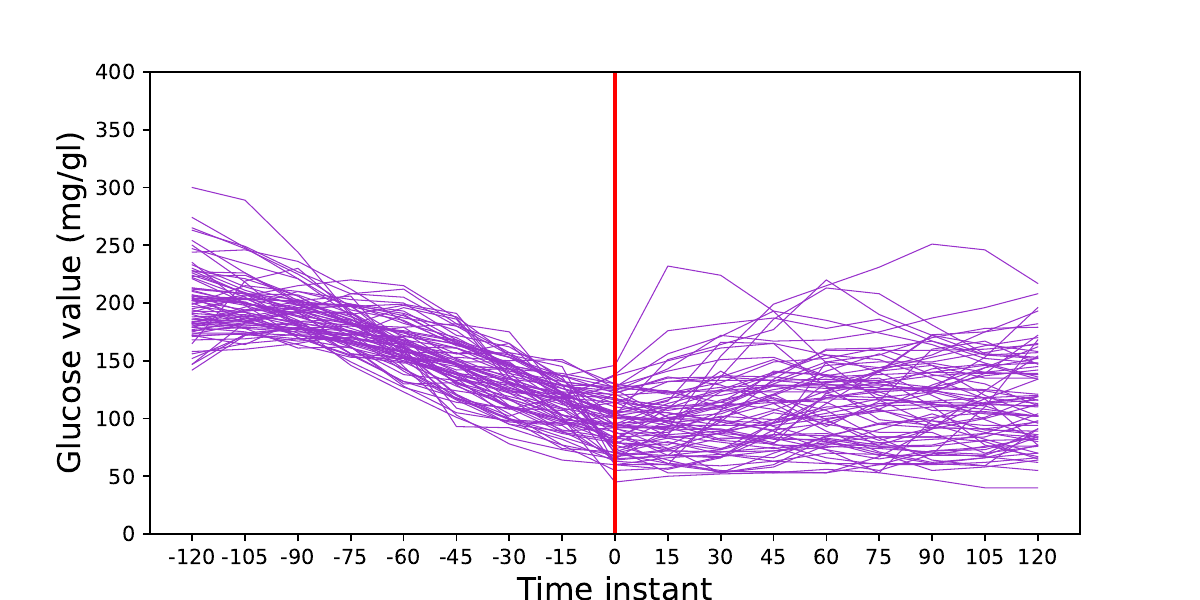}
    \caption{Cluster 10}
    \label{fig:cluster_10}
  \end{subfigure}
  \hfill
  \caption{Representation of glucose over time for clusters 4 and 10.}
\label{fig:clusters}
\end{figure}

We used k-means clustering to cluster segments based on the 2-hour window of glucose measurements before the meal.
We tried $ k \in {3, 5, 7, 9, 11,
  13, 15, 17, 19}$ and selected $k=15$ clusters where the
intra-distance leveled off, and improved only slightly for higher values of $k$. 100 random restarts where executed and the
cluster assignment with the best intra-distance was used. To give an example of the clusters obtained, Figure
\ref{fig:clusters} shows the blood glucose concentration for the segments in
Clusters 4 and 10. Each graph shows the glucose value over time. We can see how the behavior of the segments is similar for the first two hours between the segments of the respective clusters, and it is not until food intake, the red vertical line, that they begin to vary to a greater extent.

\subsection{Training, validation and Test Split}

For each one of the k clusters, a first division is made into two groups, test with one-third of the cases and, on the other hand, validation-training with the remaining two-thirds. This division will be performed only once, and the same for all the techniques in this article. The validation-training set is then divided into validation, with one-third of the cases in the subgroup and the rest for training.

In Table \ref{tab:tr_ts_split}, we show the total number of cases for each cluster and the division for training and testing.
\begin{table}[]
\centering
\resizebox{5cm}{!} {
\begin{tabular}{c|c|c|c}
\textbf{Cluster} & \textbf{Total} & \textbf{Training-validation} & \textbf{Test} \\ \hline
1                & 99             & 66                & 33            \\
2                & 92             & 61                & 31            \\
3                & 21             & 14                & 7             \\
4                & 75             & 50                & 25            \\
5                & 143            & 95                & 48            \\
6                & 137            & 91                & 46            \\
7                & 44             & 29                & 15            \\
8                & 162            & 108               & 54            \\
9                & 16             & 10                & 6             \\
10               & 75             & 50                & 25            \\
11               & 23             & 15                & 8             \\
12               & 108            & 72                & 36            \\
13               & 183            & 122               & 61            \\
14               & 171            & 114               & 57            \\
15               & 95             & 63                & 32           
\end{tabular}
}
\caption{Division of the cases of each cluster between training and test}
\label{tab:tr_ts_split}
\end{table}

\subsection{Modeling} \label{sec:modeling}
%Figure \ref{fig:Tr_val_test} shows the division of one of the clusters, cluster N, into subsets and the process of obtaining the results for  $\epsilon\text{-Lexicase ISIGE}$. First, the training phase is performed, consisting of 30 runs. We will take the best model for each run based on the mean RMSE (MRMSE) for the different segments. Through a validation stage, we will select a single model, and after the test phase, we obtain the results of  $\epsilon\text{-Lexicase ISIGE}$  for cluster N using a test stage. This process is repeated with all the clusters to obtain one model for each cluster.
The process for obtaining the results is similar to the other two techniques.

\begin{figure}[h]
    \centering
    
    \includegraphics[width=\textwidth]{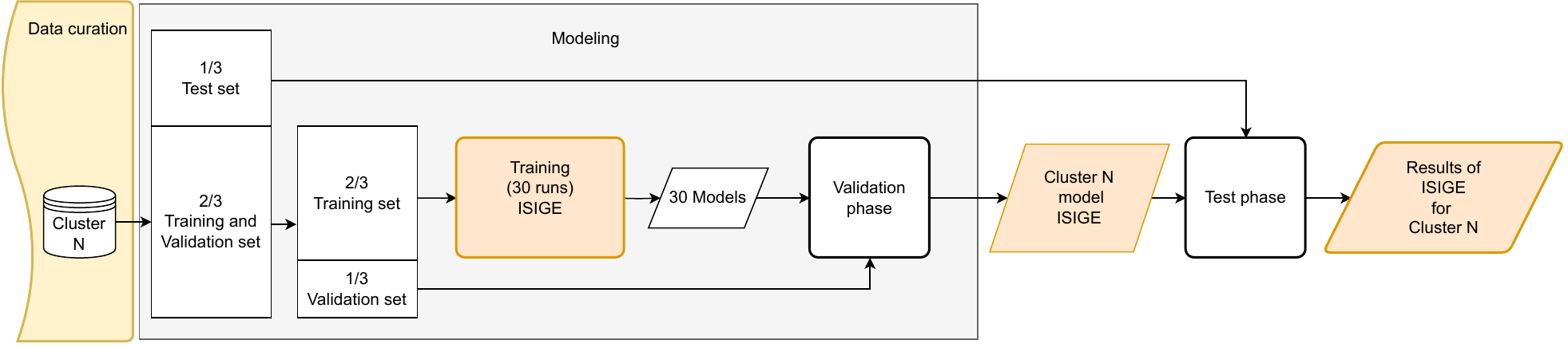}
    \caption{Generation of the results $\epsilon\text{-Lexicase ISIGE}$ for cluster N. Division of cluster N into training, validation and test sets used in the homonymous phases.}
    %caption{Generation of the $\epsilon\text{-Lexicase ISIGE}$results for cluster N. Taking cluster N belonging to the group of K clusters, three sets are divided, training, validation and test used in the homonymous phases. Next, 30 models are trained using $\epsilon\text{-Lexicase ISIGE}$, and the model of cluster N is selected through validation. After the test phase, the $\epsilon\text{-Lexicase ISIGE}$ results for cluster N are obtained.}
    \label{fig:Tr_val_test}

\end{figure}

%\color{blue}
In Figure \ref{fig:Tr_val_test}, cluster N is divided into subsets to obtain the results for ISIGE. The process consists of three stages: training, validation, and testing. During the training phase, 30 runs are performed, and the best model for each run is selected based on the mean RMSE (MRMSE) for different segments. In the validation stage, a single model is chosen, and in the testing phase, we obtain the results of the technique %$\epsilon\text{-Lexicase ISIGE}$
for cluster N. This process is repeated for all clusters to obtain a model for each cluster. 
%\color{red}
%The method for obtaining the results is similar to the other two techniques.
%The method of obtaining the results is similar in the other two techniques we have used for comparison.
%\color{black}

\begin{comment}

We implemented three different techniques for obtained prediction models:

\begin{itemize}
    \item ISIGE (section \ref{subsec:ISIGE}).
    \item SINDY (section \ref{subsec:SINDy}).
    \item ANNs.
\end{itemize}
\end{comment}

In addition to our proposal (ISIGE) and SINDy, we have experimented ANNs. As we have already stated, blood glucose prediction is an important and complex task, and ANNs have been proven valuable tools in the last years. In a recent literature review on glucose and hypoglycemia prediction methods using data from real patients \cite{FELIZARDO2021glucose_prediction}, the authors found that ANNs, particularly recurrent neural networks (RNN) and hybrid models, performed well in predicting glucose levels. Moreover, many software tools in different languages are available on the market, and their use is increasingly ubiquitous in all kinds of problems, making them a recommendable option for the problem of blood glucose level prediction.
In another study \cite{Tena2021BGpredNN}, the performances of several neural network models were compared for predicting blood glucose in patients with diabetes. Following the findings of the previous study, we have chosen, for this work, three ANNS models: Meijner~\cite{meijner2017blood}, Mirshekarian~\cite{mirshekarian2017LSTMs}, and Sun~\cite{sun2018NN}.

Based on our experimental findings and comparative analysis, it was determined that the ANN version developed by Mirshekarian exhibited superior performance. Therefore, to clearly represent ANNs, only the outcomes for the model of Mirshekarian will be showcased for comparison with SINDy and ISIGE.
\color{black}

\section{Experimental Results} \label{sec:results}
The methods studied in this work are ANNs, particularly of the Mirshekarian type, SINDy, and ISIGE, and their performance is compared with a simple baseline that is calculated by taking the mean of the glucose values at each time step over the segments of the training set.

\subsection{MRMSE Results} \label{cap:MRMSE_Errors}

Table~\ref{tab:results-rmse} shows the MRMSE in the test for each method and all clusters.

The MRMSE values of SINDy and  SGE are lower than the
values for the mean prediction in most cases. The average MRMSE in all clusters is 38.96 [mg/L] and 37.79 [mg/L], respectively. The best
values for each cluster and the best mean are marked.

\begin{table}[h]
\centering
\begin{tabular}{cccccc}
         % Adj. to test size
         % Data was changed on 2/1/2023.
         % Mean baseline calculated for training set and evaluated for test set (21/2/2023)
Cluster ID & \# Segments & Mean pred.                      & ANN      & SINDy                           &  ISIGE       \\ \hline
$1 $       & $99  $      & $42.14$                         & $45.67$ & $52.52$                         & \textbf{41.17} \\
$2 $       & $92  $      & $38.85$                         & $43.56$ & $35.45$                         & \textbf{32.81} \\
$3 $       & $21  $      & \textbf{50.25}                  & $52.58$ & $55.65$                         & $55.03$                         \\
$4 $       & $75  $      & $43.63$                         & $47.05$ & $44.42$                         & \textbf{42.87} \\
$5 $       & $143 $      & $30.15$                         & $32.14$ & $28.73$                         & \textbf{26.43} \\
$6 $       & $137 $      & $24.75$                         & $26.96$ & \textbf{22.42}                  & 23.04                           \\
$7 $       & $44  $      & $53.84$                         & $62.53$ & \textbf{44.52}                  & 48.14                           \\
$8 $       & $162 $      & $37.20$                         & $38.73$ & \textbf{30.78}                  & 32.44                           \\
$9 $       & $16  $      & $71.29$                         & $81.12$ & \textbf{44.82}                  & 45.98                           \\
$10$       & $75  $      & $36.34$                         & $36.44$ & $30.22$                         & \textbf{29.92} \\
$11$       & $23  $      & \textbf{43.50}                  & $46.08$ & $45.87$                         & 50.48                           \\
$12$       & $108 $      & $53.33$                         & $53.17$ & $50.27$                         & \textbf{48.65} \\
$13$       & $183 $      & $36.17$                         & $37.39$ & $34.81$                         & \textbf{34.18} \\
$14$       & $171 $      & $32.95$                         & $34.94$ & $31.50$                         & \textbf{30.97} \\
$15$       & $95  $      & $27.08$                         & $30.34$ & $32.41$                         & \textbf{24.72} \\
Avg.       & $96.27$      & $41.43$                         & $44.58$ & $38.96$                         & $37.79$  
\end{tabular}

\caption{MRMSE (mg/L) for the test set for all methods and clusters.}
\label{tab:results-rmse}
\end{table}

\subsection{Parkes Error Grid Analysis} \label{cap:Parkes_Errors}

To measure the error of these techniques, we should not only look
at the MRMSE. Since we are dealing with a clinical problem, it is
necessary to take into account the clinical consequences.
An established method for assessing errors made by blood glucose estimation or prediction systems is the Parkes Error Grid (PEG), \cite{parkes2000PEG}. In the PEG the actual value of blood glucose and the predicted values are plotted on a grid and associated with risk levels. These levels go from A to E and are described in Table~\ref{tab:PEG-zones}.
\begin{table}
\centering
  \begin{tabular}{|c|l|}
\hline
Zone & Description                                                              \\ \hline
A    & The actual value and the predicted value are similar.                    \\ \hline
B    & Does not lead to any action by the patient or leads to benign treatment. \\ \hline
C    & Overcorrection of acceptable blood glucose levels                        \\ \hline
D    & Dangerous failure to detect and treat errors                             \\ \hline
E    & Erroneous treatment, contradictory to that actually needed               \\ \hline
\end{tabular}
  \caption{Parkes error grid zones.}\label{tab:PEG-zones}
\end{table}

For this work we use the version of PEG designed for type-I
diabetes~\cite{parkes2000PEG}.

 Table \ref{tab:parkes_err} shows the percentage of cases in
each of the zones of the PEG, for each method and divided by
clusters.
Zone E has been omitted, since its value was zero for all methods in all clusters. Analyzing the results, most of the points are in zones A and B in all the methods, with SINDy standing out as the method with the highest mean number of points in zone A, 59.0\%, and with a total of 92.4\% when adding the means of zones A and B. In addition, we must also highlight the results obtained by  ISIGE, achieving 93.6 \% in total for the sum of the averages of zones A and B. As we move away from zones A and B, the
predictions made put the patient at greater risk, with C being not
very recommendable, D potentially dangerous, and E leading to an
erroneous and extremely dangerous treatment. In this aspect, the
models have an average percentage below 10\% in zone C, with
SGE standing out with only 6.2\%. In the case of
zone D, SINDy and ISIGE, are below
1.0\%, with the lowest being ISIGE, with an average of only 0.2\% of the points in this zone.

\begin{table}[]
\resizebox{\textwidth}{!}{
\centering
\begin{tabular}{cc|cccc|cccc|cccc|cccc}
                             &               & \multicolumn{4}{c|}{Mean prediction} & \multicolumn{4}{c|}{Mirshekarian (ANN)} & \multicolumn{4}{c|}{SINDy}   & \multicolumn{4}{c}{ISIGE} \\ \hline
\multicolumn{1}{c|}{Cluster} & Test segments & A       & B       & C       & D      & A        & B        & C       & D      & A     & B     & C     & D    & A            & B           & C           & D          \\ \hline
\multicolumn{1}{c|}{1}       & 33            & 48.5    & 42.4    & 8.1     & 1      & 43.2    & 46.6    & 9.1    & 1.1   & 45.5 & 39.8 & 14.8 & 0    & 48.5         & 42.8        & 8.7         & 0          \\
\multicolumn{1}{c|}{2}       & 31            & 55.2    & 40.9    & 2.9     & 1.1    & 48.8    & 45.6    & 3.6    & 2.0   & 58.1 & 38.3 & 3.2  & 0.4 & 59.7         & 35.9        & 4.4         & 0          \\
\multicolumn{1}{c|}{3}       & 7             & 76.2    & 23.8    & 0       & 0      & 60.7    & 35.7    & 3.6    & 0     & 76.8 & 14.3 & 8.9  & 0    & 60.7         & 39.3        & 0           & 0          \\
\multicolumn{1}{c|}{4}       & 25            & 51.6    & 36.9    & 8.9     & 2.7    & 43.0    & 43.5    & 11.0   & 2.5   & 43.0 & 42.5 & 13.0 & 1.5 & 50.5         & 38.0        & 10.5        & 1          \\
\multicolumn{1}{c|}{5}       & 48            & 57.2    & 30.3    & 12.5    & 0      & 50.0    & 34.9    & 14.6   & 0.5   & 60.7 & 35.4 & 3.9  & 0    & 64.1         & 31.0        & 4.9         & 0          \\
\multicolumn{1}{c|}{6}       & 46            & 61.6    & 33.1    & 5.3     & 0      & 61.7    & 33.7    & 4.6    & 0     & 70.1 & 25.5 & 4.3  & 0    & 71.2         & 26.9        & 1.9         & 0          \\
\multicolumn{1}{c|}{7}       & 15            & 67.4    & 28.1    & 3.7     & 0.7    & 52.5    & 40.8    & 5.8    & 0.8   & 70.8 & 25.0 & 3.3  & 0.8 & 66.7         & 27.5        & 5.0         & 0.8        \\
\multicolumn{1}{c|}{8}       & 54            & 51      & 40.1    & 8.8     & 0      & 47.2    & 43.3    & 9.5    & 0     & 59.3 & 36.3 & 4.4  & 0    & 56.7         & 36.3        & 6.9         & 0          \\
\multicolumn{1}{c|}{9}       & 6             & 59.3    & 27.8    & 7.4     & 5.6    & 41.7    & 45.8    & 6.2    & 6.2   & 62.5 & 31.3 & 2.1  & 4.2 & 62.5         & 31.2        & 6.2         & 0          \\
\multicolumn{1}{c|}{10}      & 25            & 42.2    & 45.8    & 11.1    & 0.9    & 44.0    & 42.5    & 13.0   & 0.5   & 63.5 & 27.0 & 8.0  & 1.5 & 56.5         & 35.0        & 8.5         & 0          \\
\multicolumn{1}{c|}{11}      & 8             & 52.8    & 38.9    & 6.9     & 1.4    & 42.2    & 51.6    & 4.7    & 1.6   & 54.7 & 40.6 & 4.7  & 0    & 50.0         & 34.4        & 15.6        & 0          \\
\multicolumn{1}{c|}{12}      & 36            & 46.9    & 40.1    & 9.6     & 3.4    & 43.8    & 39.9    & 12.5   & 3.8   & 46.9 & 43.8 & 7.3  & 2.1 & 48.6         & 41.7        & 8.0         & 1.7        \\
\multicolumn{1}{c|}{13}      & 61            & 56.8    & 34.6    & 8.6     & 0      & 52.3    & 38.3    & 9.4    & 0     & 55.3 & 36.1 & 8.4  & 0.2 & 59.2         & 33.4        & 7.2         & 0.2        \\
\multicolumn{1}{c|}{14}      & 57            & 59.8    & 36.1    & 4.1     & 0      & 55.9    & 39.3    & 4.6    & 0.2   & 63.2 & 33.3 & 3.5  & 0    & 62.9         & 34.2        & 2.9         & 0          \\
\multicolumn{1}{c|}{15}      & 32            & 61.1    & 30.2    & 8.7     & 0      & 57.8    & 30.9    & 11.3   & 0     & 55.1 & 32.0 & 11.3 & 1.6 & 65.2         & 32.8        & 2.0         & 0          \\ \hline
\multicolumn{1}{c|}{Mean}    & 32.3         & 56.5    & 35.3    & 7.1     & 1.1    & 49.7    & 40.8    & 8.2   & 1.3   & 59.0 & 33.4 & 6.7  & 0.8 & 58.9         & 34.7        & 6.2         & 0.2       
\end{tabular}
}
\caption{Percentage of prediction / measurement pairs in the Parkes error zones by cluster and method}
\label{tab:parkes_err}
\end{table}

\begin{figure}
\centering
  
  \begin{subfigure}{0.49\textwidth}
    \includegraphics[width=1 \textwidth]{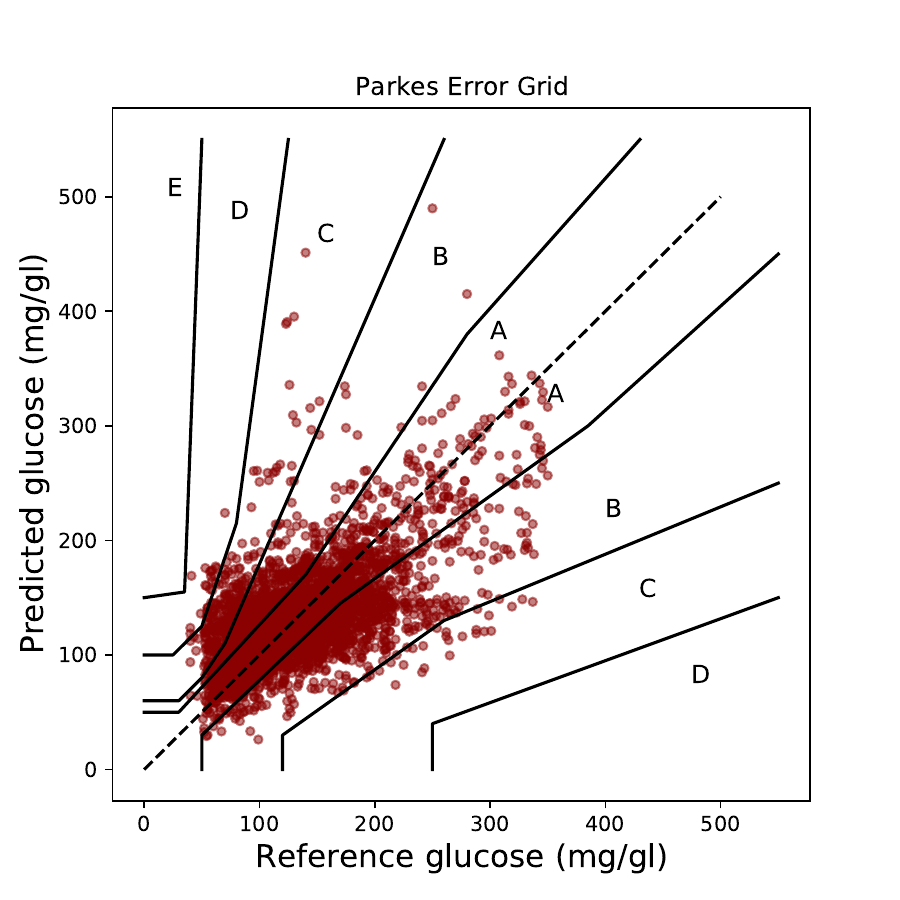}
    \caption{SINDy}
    \label{fig:PEG_SINDy}
  \end{subfigure}
  \hfill
  \begin{subfigure}{0.49\textwidth}
    \includegraphics[width=1 \textwidth]{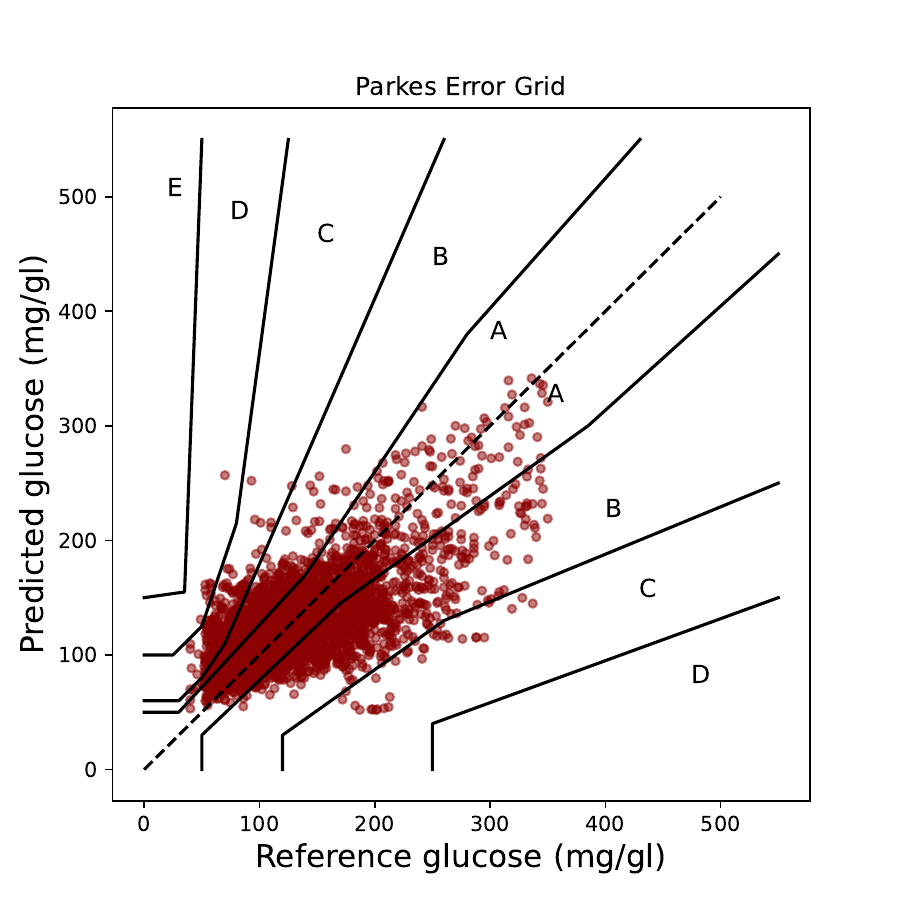}
    \caption{ISIGE}
    \label{fig:PEG_lexicase}
  \end{subfigure}
  \hfill
  \begin{subfigure}{0.49\textwidth}
    \includegraphics[width=1 \textwidth]{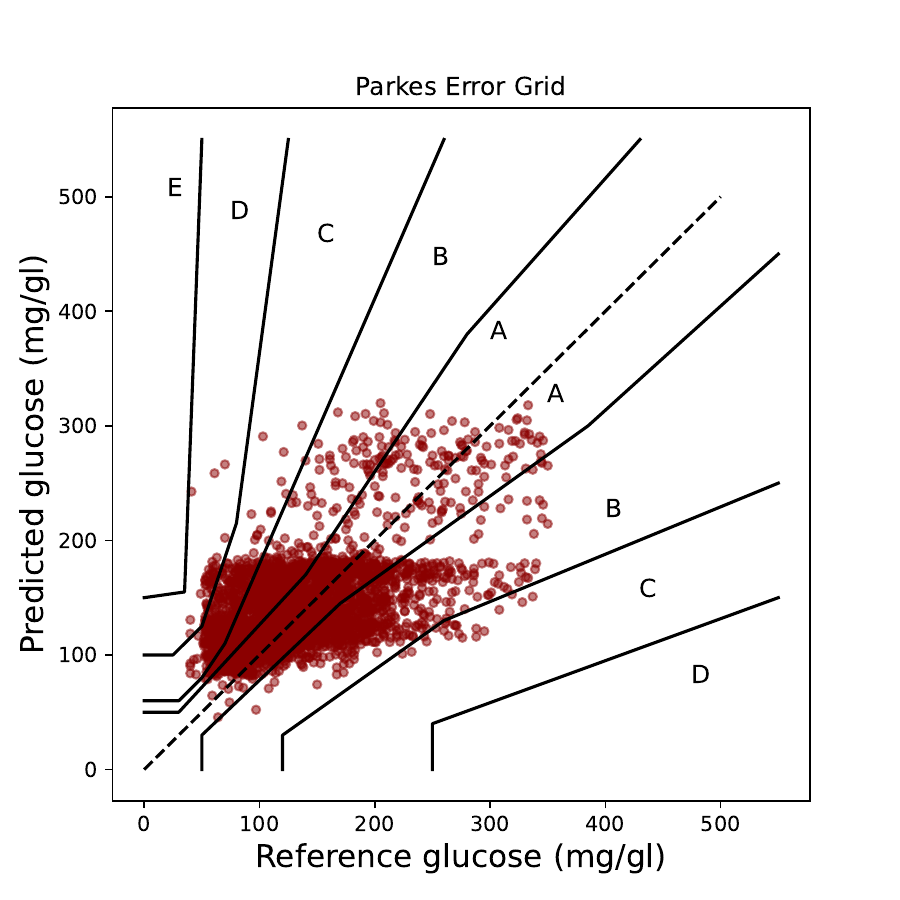}
    \caption{Mirshekarian}
    \label{fig:PEG_Mirshekarian}
  \end{subfigure}
  \hfill
  \caption{Plot showing the PEG for each method and all clusters.}
\label{fig:PEG_all}
\end{figure}

Figure \ref{fig:PEG_all} shows the PEG error zones for all the
clusters and all methods. We can see that most of the points are
located between zones A and B. To a minor amount, we can find points
in zones C and D, particularly in the left zone of the grid. This zone
is related to the actual values associated with hypoglycemia (values
below 70), those predictions that give a high glucose value being in
this zone are potentially dangerous, so it is of great interest to
avoid these situations. On the other hand, it is easier to detect
hypoglycemic values compared to hyperglycemic ones.

%%%%%%%%%
%%%%%%%%%

\begin{figure}  
\centering
  \begin{subfigure}{0.49\textwidth}
    \includegraphics[width=1 \textwidth]{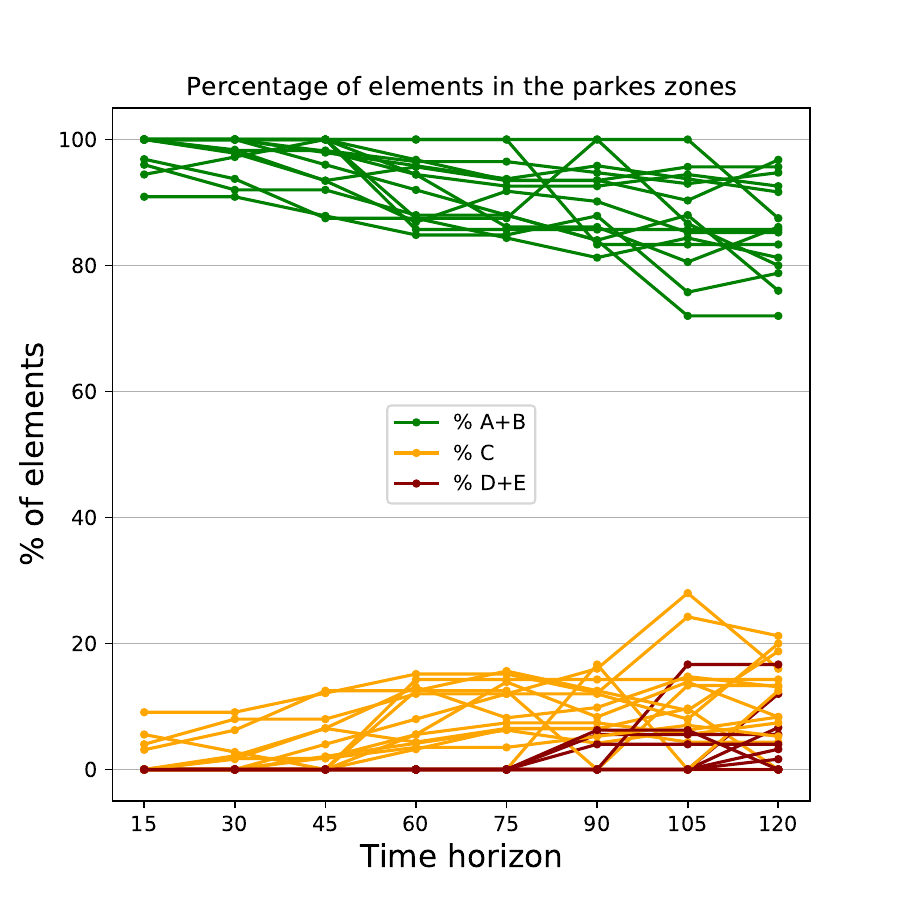}
    \caption{SINDy}
    \label{fig:parkes_plot_th_all_SINDy}
  \end{subfigure}
  \hfill
  \begin{subfigure}{0.49\textwidth}
    \includegraphics[width=1 \textwidth]{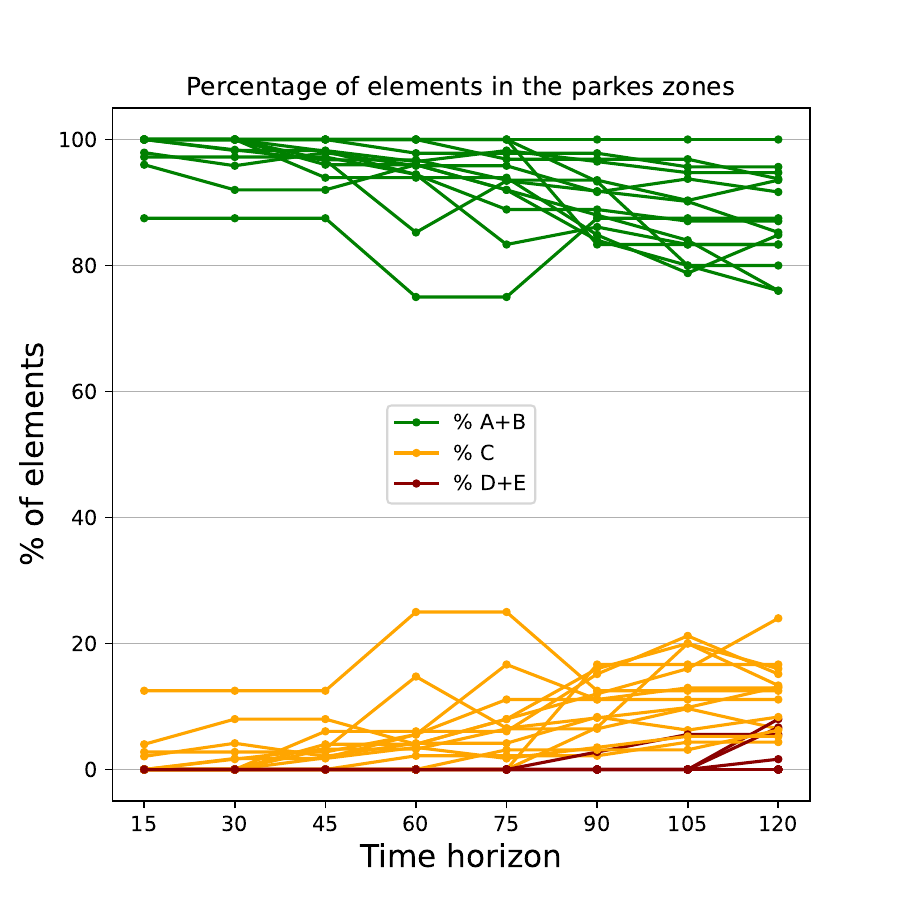}
    \caption{ISIGE}
    \label{fig:parkes_plot_th_all_lexicase}
  \end{subfigure}
  \hfill
  \begin{subfigure}{0.49\textwidth}
    \includegraphics[width=1 \textwidth]{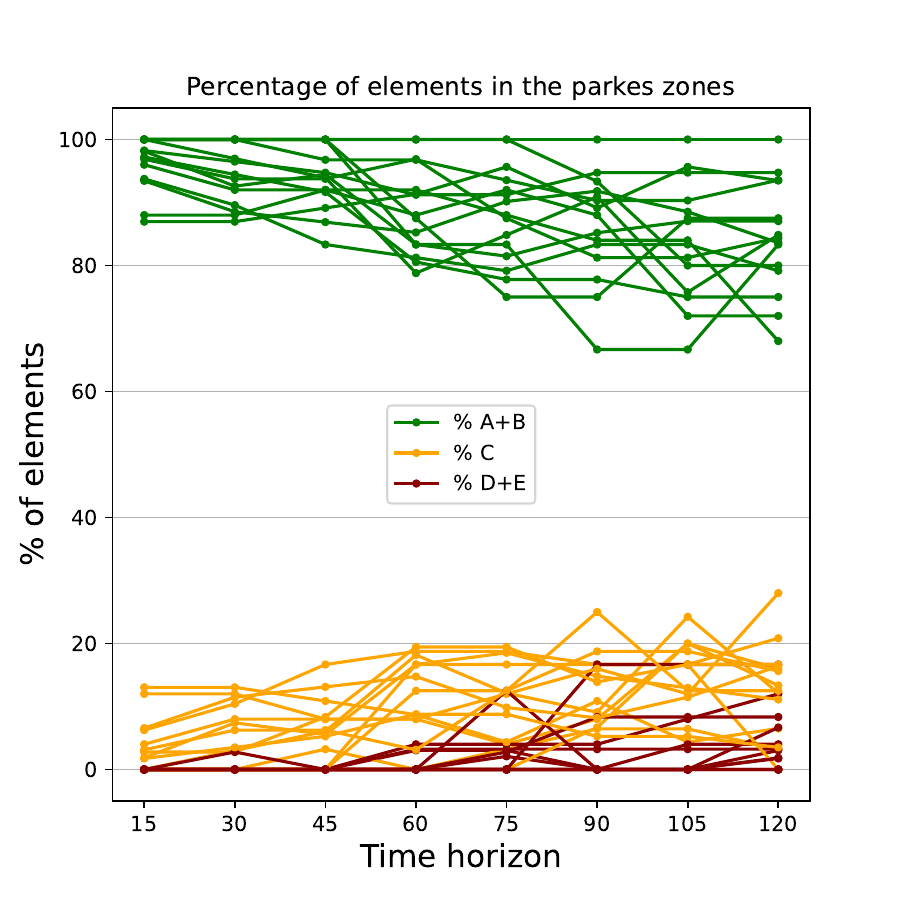}
    \caption{Mirshekarian}
    \label{fig:parkes_plot_th_all_Mirshekarian}
  \end{subfigure}
  \hfill
  \caption{Plot showing the percentage of elements in the different Parkes Error Grid zones for each method and all clusters.}
  \label{fig:parkes_plot_th_all}
\end{figure}

One of the advantages of using FDEs is that we can obtain forecasting for several time steps. In this work, eight steps predictions are made for each
segment,(i.e. forecast horizon of 2 hours in 15-minute steps).
In Figure \ref{fig:parkes_plot_th_all} we show the distribution of points
in the different zones of the PEG. Three divisions have been made
by joining zones of the PEG, A + B (green), C (orange), and D + E (dark red).
The x-axis
shows the time horizon, and the y-axis shows the percentage
of glucose prediction and measurement pairs in each zone.

By comparing the percentage of cases in zones A and B obtained by each of the techniques for each time horizon, we can see that the lowest value is around 80\% for our proposal, 70\% for SYNDy and 60\% for Mirshekarian.
%%%%%%%%%

%%%%%%%%%

In Figure
\ref{fig:parkes_plot_th_all_lexicase} for SGE, most points are concentrated in zones A and B at the beginning.  The further we move away from the initial time instant, the more points appear in zone C and finally reach zone D. This behavior can be explained due to the increasing
difficulty of the prediction for longer forecast periods.

Currently, predictions are made at 15 minutes step, but applying the same methodology makes it possible to modify this interval depending on the frequency with which the data have been collected on CGM system. 
The Parkes error Grid of each time step for
all methods is attached in \ref{aped.A}.
\newpage
\subsection{Expressions and interpretability} \label{cap:expressions}

The interpretability of machine learning models is a topic of significant interest, yet a consensus on its definition, quantification, and measurement remains elusive. Complexity, transparency, and the ability to be simulated are among the key characteristics often associated with interpretability. However, interpretations may vary depending on the perspective of the evaluator, with mathematicians and healthcare professionals likely to have differing opinions on the same model. In this study, focusing on the healthcare domain, we specifically examine the transparency of models as a crucial aspect of interpretability. Based on the work of Lipton (2018)\cite{Lipton2018} and Belle (2021) \cite{Belle2021Explainable},  three dimensions are proposed to analyze model transparency: simulatability, decomposability, and algorithmic transparency.

\begin{itemize}
    \item Simulatability refers to the capacity of the model to be simulated by a human, with simplicity and compactness being key attributes. Simple and compact models are more likely to fall into this category. However, it is essential to note that more than simplicity alone is required, as an excessive number of simple rules can hinder the ability of a human to mentally calculate the decisions of the model. Conversely, some complex models, such as neural networks without hidden layers, may still exhibit simulatability.
    
    \item Decomposability entails breaking down the model into interpretable parts, including inputs, parameters, and computations, and subsequently explaining each component. Unfortunately, not all models satisfy this property, making it challenging to explain their inner workings comprehensively.
    
    \item Algorithmic transparency focuses on understanding the procedural steps employed by the model to generate its outputs. Models that employ clear procedures, such as similarity-based classifiers like K-nearest neighbours, exhibit algorithmic transparency. On the other hand, complex models like neural networks construct elusive loss functions, and the solution to the training objective often requires approximation. Inspecting the model through mathematical analysis is the primary requirement for it to fall into this category.
\end{itemize}

Some models inherently possess one or more of these qualities, making them candidates for transparent models. For example, logistic/linear regression models used to predict continuous and categorical targets, where the target is a linear combination of variables, can be considered transparent due to their clear modelling choices. However, to maintain transparency, such models must be limited, and the variables used should be understandable to the intended users. The complexity of the model is directly linked to its transparency. Even if a model satisfies several transparency dimensions, a highly complex procedure or a large number of dimensions can render the model less interpretable.

The definition of transparency by Lipton and Belle aligns with the models obtained by SINDy and ISIGE, suggesting their transparency and explainability. Evaluating the complexity of these models confirms their interpretability, although SINDy models with numerous terms may pose challenges in interpretation due to the increased complexity introduced.

Table \ref{tab:Expression_eLexicase_ISIGE_2} presents the expressions obtained by ISIGE for the different clusters. Notably, a consistent element observed across all expressions is the inclusion of $G(t_n)$ at the beginning. This common occurrence arises from the grammar utilized during the expression generation process, ensuring that all the expressions obtained conform to the form of FDEs. By imposing this starting point, the resulting expressions adhere to the desired mathematical structure.

To facilitate comprehension of the expressions, they have been simplified using the Python sympy library \cite{meurer2017sympy}, which enables a more accessible understanding of the obtained mathematical representations. The employment of this library aids in unraveling the complexity of the expressions and enhances their interpretability.

%Of particular significance is the recurring presence of the variable HR, denoting heart rate. Extensive research, as demonstrated in prior studies (Rothberg et al., 2016)~\cite{rothberg2016heart_rate_Glucose}, has established the correlation between heart rate and blood glucose levels. Therefore, the inclusion of HR in the expressions is justified, considering its established connection with the target variable. Furthermore, it is worth noting that HR frequently appears in conjunction with carbohydrate intake, indicating a certain frequency of occurrence. This observation could potentially warrant further investigation in future studies, as it may hold relevance and provide valuable insights into the relationship between heart rate, carbohydrate intake, and the target variable.

%\color{blue}    (heart rate)
The recurrent occurrence of the variable HR  in the expressions is interesting. Previous studies (Rothberg et al., 2016)~\cite{rothberg2016heart_rate_Glucose}, have extensively investigated the correlation between heart rate and blood glucose levels, establishing a solid connection between these variables. Therefore, the presence of HR in the expressions is justified, considering its established relationship with the target variable.
Additionally, it is worth noting that HR often co-occurs with carbohydrate intake in the expressions. This frequent association suggests a potential connection between heart rate, carbohydrate intake, and the target variable. This observation raises intriguing possibilities for future research, as it could provide valuable insights into the relationship among these variables. Further investigations in this direction may be warranted to explore and uncover the significance of this relationship.
Furthermore, it is interesting to highlight that in the solution of cluster number 8, the expression includes terms for carbohydrate intake and HR, while the term for insulin bolus is absent. This solution demonstrates satisfactory performance due to the time series behaviour of the cluster. Nevertheless, this finding calls for specialized clinicians to conduct in-depth studies and assess the validity and applicability of such a solution. However, thanks to the transparency of the model, a healthcare specialist could choose to replace this expression with another one that obtained a good result in the validation phase.% Further scrutiny and evaluation are necessary to determine its effectiveness and potential suitability in practical clinical settings.
%\color{black}

\begin{table}[h]
\resizebox{\textwidth}{!}{
\begin{tabular}{cl}
Cluster & \multicolumn{1}{c}{Expression} 
                                                                                                                                        \\
1       & $G(t_{n+1})= G(t_n) - I_B(t_n) \cdot C(t_n) + I_B(t_n) - F_{ch}(t_n) \cdot G(t_n) + 4.6$	\\
2       & $G(t_{n+1})= G(t_n) + I_B(t_n) \cdot 	\text{HR}(t_n) - G(t_n)^2/4000 + 1/(I_B(t_n) \cdot 	\text{HR}(t_n))$	\\
3       & $G(t_{n+1})= G(t_n) - 49 \cdot I_B(t_n) \cdot 	\text{HR}(t_n)/10 + F_{ch}(t_n) \cdot G(t_n) - F_{ch}(t_n) \cdot S(t_n) + 1/(F_{ch}(t_n)^2 \cdot G(t_n)^2)$	\\
4       & $G(t_{n+1})= G(t_n) - B_I(t_n) \cdot C(t_n) - 2 \cdot F_{ch}(t_n) \cdot G(t_n) + F_{ch}(t_n) \cdot 	\text{HR}(t_n) + 7.8$	\\
5       & $G(t_{n+1})= G(t_n) - B_I(t_n) \cdot G(t_n) + B_I(t_n) \cdot 	\text{HR}(t_n) - I_B(t_n) \cdot 	\text{HR}(t_n) +  9.9$	\\
6       & $G(t_{n+1})= G(t_n) +2 \cdot I_B(t_n) \cdot C(t_n) - I_B(t_n) \cdot G(t_n) - F_{ch}(t_n) \cdot G(t_n) +  8$	\\
7       & $G(t_{n+1})= G(t_n) - I_B(t_n) \cdot S(t_n)  + 49/5 - G(t_n)/C(t_n)$	\\
8       & $G(t_{n+1})= G(t_n) -F_{ch}(t_n) \cdot 	\text{HR}(t_n)  + 3.6$	\\
9       & $G(t_{n+1})= G(t_n) -C(t_n) \cdot F_{ch}(t_n) \cdot (-5 \cdot I_B(t_n) \cdot F_{ch}(t_n) \cdot 	\text{HR}(t_n)^2 + 5 \cdot F_{ch}(t_n) + 91)/5$	\\
10      & $G(t_{n+1})= G(t_n) - I_B(t_n) \cdot G(t_n) + F_{ch}(t_n) \cdot 	\text{HR}(t_n)  + 3.9$	\\
11      & $G(t_{n+1})= G(t_n) + B_I(t_n) \cdot C(t_n) - I_B(t_n) \cdot C(t_n) - I_B(t_n) \cdot G(t_n)  - 1/(I_B(t_n) \cdot 	\text{HR}(t_n))$	\\
12      & $G(t_{n+1})= G(t_n) + B_I(t_n) \cdot C(t_n) - I_B(t_n) \cdot C(t_n) - I_B(t_n) \cdot G(t_n) - F_{ch}(t_n) \cdot G(t_n) +  2.4$	\\
13      & $G(t_{n+1})= G(t_n) - I_B(t_n) \cdot G(t_n) - F_{ch}(t_n) \cdot G(t_n) + F_{ch}(t_n) \cdot 	\text{HR}(t_n)  + 9.49989$	\\
14      & $G(t_{n+1})= G(t_n) - I_B(t_n) \cdot C(t_n) + C(t_n) \cdot F_{ch}(t_n) - F_{ch}(t_n) \cdot G(t_n)  + 5.9$ 	\\
15      & $G(t_{n+1})= G(t_n) - B_I(t_n) \cdot G(t_n) + B_I(t_n) \cdot 	\text{HR}(t_n) - I_B(t_n) \cdot 	\text{HR}(t_n)  + 7.3027$
\end{tabular}
}
\caption{Expressions for each cluster using ISIGE}
\label{tab:Expression_eLexicase_ISIGE_2}
\end{table}
In this section we only discuss the expressions obtained by ISIGE, but in Appendix \ref{tab:Expression_SINDy} we attach those obtained by SINDy.

\section{Conclusion and future work} \label{sec:conclusion}

In conclusion, this paper proposes a novel approach to glucose prediction in diabetes management that emphasises interpretability: Interpretable Sparse Identification by Grammatical Evolution (ISIGE). The proposed technique combines clustering with ISIGE to obtain finite difference equations that predict postprandial glucose levels up to two hours after meals.

In this study, we have employed data from 24 different participants with diabetes mellitus type-I. The data were divided into four-hour segments with two hours before and two hours after a meal, i.e. carbohydrate intake. Clustering was performed based on the blood glucose values for the two-hour window before the meal, dividing the cases into 15 clusters. Forecasts were calculated for the two-hour window after the meal. The results of the newly proposed method are compared to SINDy and artificial neural networks (ANN).
  
Using Parkes Error Grid to quantify the safety and robustness of the predictions, the essential conclusions of this work are:

\begin{itemize}
    \item Predictions produced by comparison methods (SINDy and ANNs) are in zones A and B, indicating that the predictions of these models are generally safe. ISIGE achieves 93.6 \% predictions in  zones A and B, which is the highest among all methods.

    \item SINDy has the highest number of predictions (59 \%) in zone A, and 92.4 \% in either zone A or B.

    \item All models have fewer than 10 \% of predictions in zone C, with ISIGE standing out with only 6.2 \%. SINDy and ISIGE have below 1.0 \% of predictions in zone D, with the lowest being ISIGE, with an average of only 0.2 \%. None of the studied methods has predictions in zone E.

    \item Although only few predictions are in zones C and D, they represent a risk for participants and should be monitored.
    
    \end{itemize}
    
We have used the Mean root mean square error (MRMSE) to quantify the accuracy of the predictions. In this sense, the conclusions are:

\begin{itemize}
    \item The average MRMSE over all clusters is 38.96 [mg/L] for SINDy and 37.79 [mg/L] ISIGE. The results indicate that the models have different performance levels across different clusters. 
    \item Although ANN is one of the best-performing techniques for glucose prediction and one of the most commonly used methods in this field, in our study, both SINDy and ISIGE obtained better results in most clusters.
\end{itemize}

%This paper highlights the importance of interpretability in glucose prediction models and how it can impact clinical decision-making. We conclude that ISIGE can potentially assist diabetes clinicians in developing personalised treatment plans for their patients, offering a promising alternative to traditional black-box models like ANNs. 

%\color{blue}
Transparency plays a pivotal role in enhancing the interpretability of machine learning models in the healthcare domain. By focusing on the dimensions of simulatability, decomposability, and algorithmic transparency, we can assess and analyze the transparency of models. Transparent models exhibit simplicity, compactness, and transparent procedures, making them more amenable to human understanding. The insights gained from this study provide valuable guidance for future developments in interpretability, helping to advance the explainability of machine learning models in healthcare and beyond.  We conclude that ISIGE can potentially assist diabetes clinicians in developing personalised treatment plans for their patients, offering a promising alternative to traditional black-box models like ANNs. 
%\color{black}

%In the future, the most critical step to complete the functioning of our proposal is developing a method to classify the new data within the clusters for the subsequent application of the prediction model.

%In addition, it would be interesting to carry out a more exhaustive analysis of the evolution of the predictions in the different time horizons and their correlation with the pre-meal behaviour of each cluster.

%%%%%%%%%%%%%%%%%%%%%%%%%%%%%%%%%%%%%%%%

\appendix

\section*{Acknowledgements}
This work has been supported by The Spanish Ministerio de Innovaci\'on Ciencia y Universidad -grants PID2021-125549OB-I00, PDC2022-133429-I00 and RTI2018-095180-B-I00.
\bibliographystyle{elsarticle-num}
\bibliography{Glucose_Prediction}

\newpage
\appendix
\section{Parkes Error Grid per time horizon}
\label{aped.A}
\begin{figure}

\centering
  \begin{subfigure}{0.49\textwidth}
    \includegraphics[width=\textwidth]{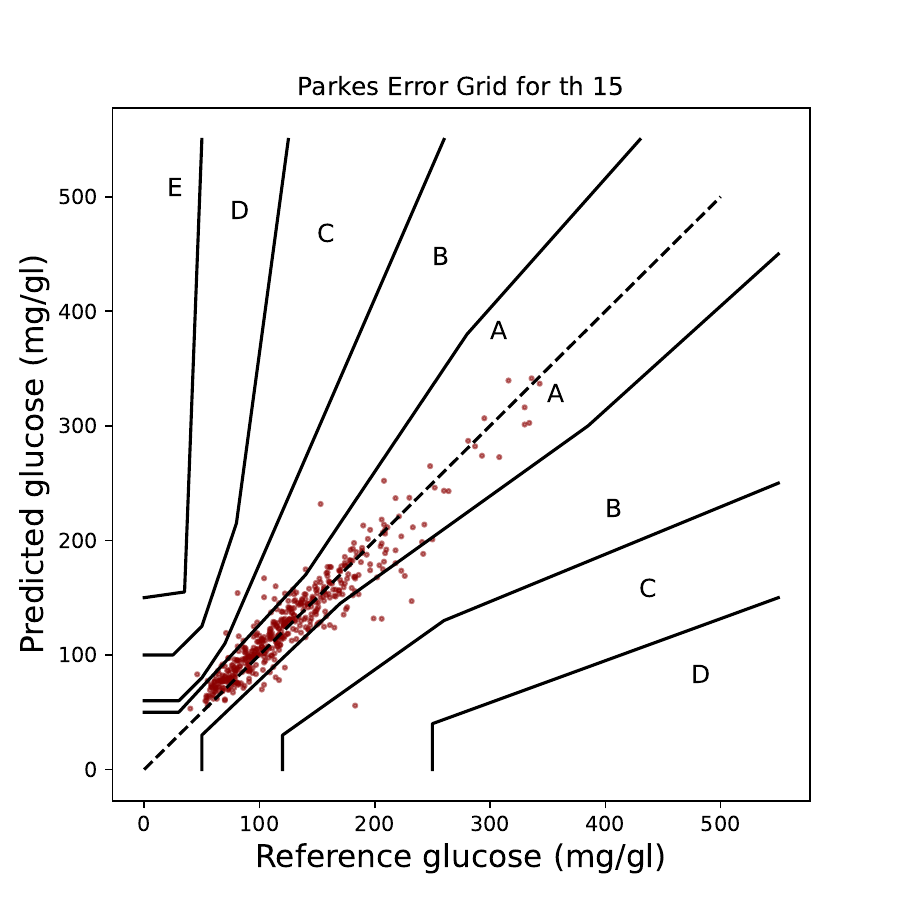}
    \caption{Time horizon 15}
    \label{fig:parkes_th_15_LexGE}
  \end{subfigure}
  \hfill
  \begin{subfigure}{0.49\textwidth}
    \includegraphics[width=\textwidth]{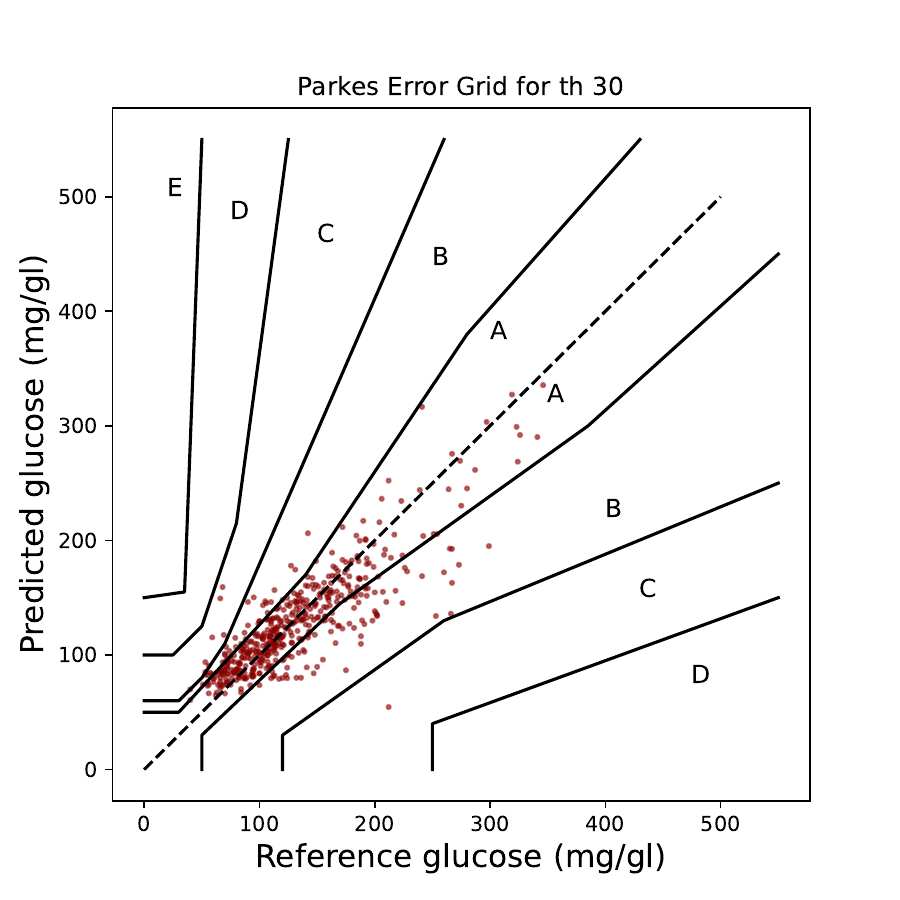}
    \caption{Time horizon 30}
    \label{fig:parkes_th_30_LexGE}
  \end{subfigure}
  \hfill
  \begin{subfigure}{0.49\textwidth}
    \includegraphics[width=\textwidth]{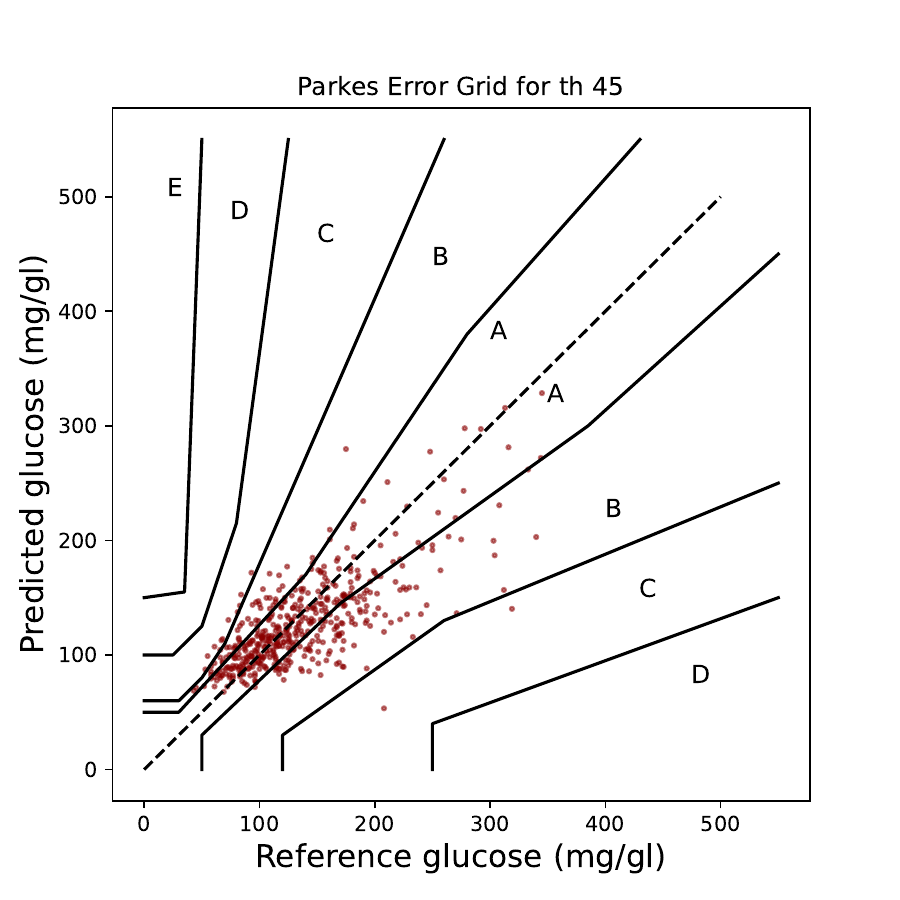}
    \caption{Time horizon 45}
    \label{fig:parkes_th_45_LexGE}
  \end{subfigure}
  \hfill
  \begin{subfigure}{0.49\textwidth}
    \includegraphics[width=\textwidth]{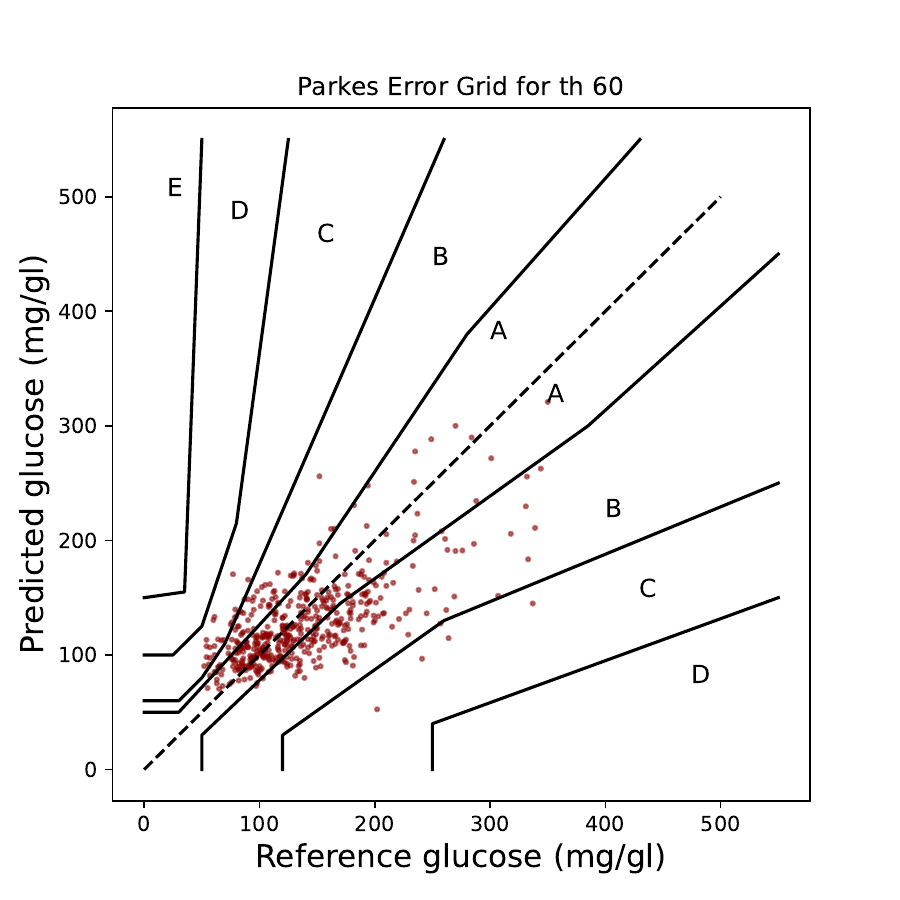}
    \caption{Time horizon 60}
    \label{fig:parkes_th_60_LexGE}
  \end{subfigure}
  \caption{Parkes Errors Grid by Time Horizon for ISIGE, all clusters (part 1).}
  \label{fig:parkes_th_split_LexGE}
\end{figure}

\begin{figure}

\centering
  \begin{subfigure}{0.49\textwidth}
    \includegraphics[width=\textwidth]{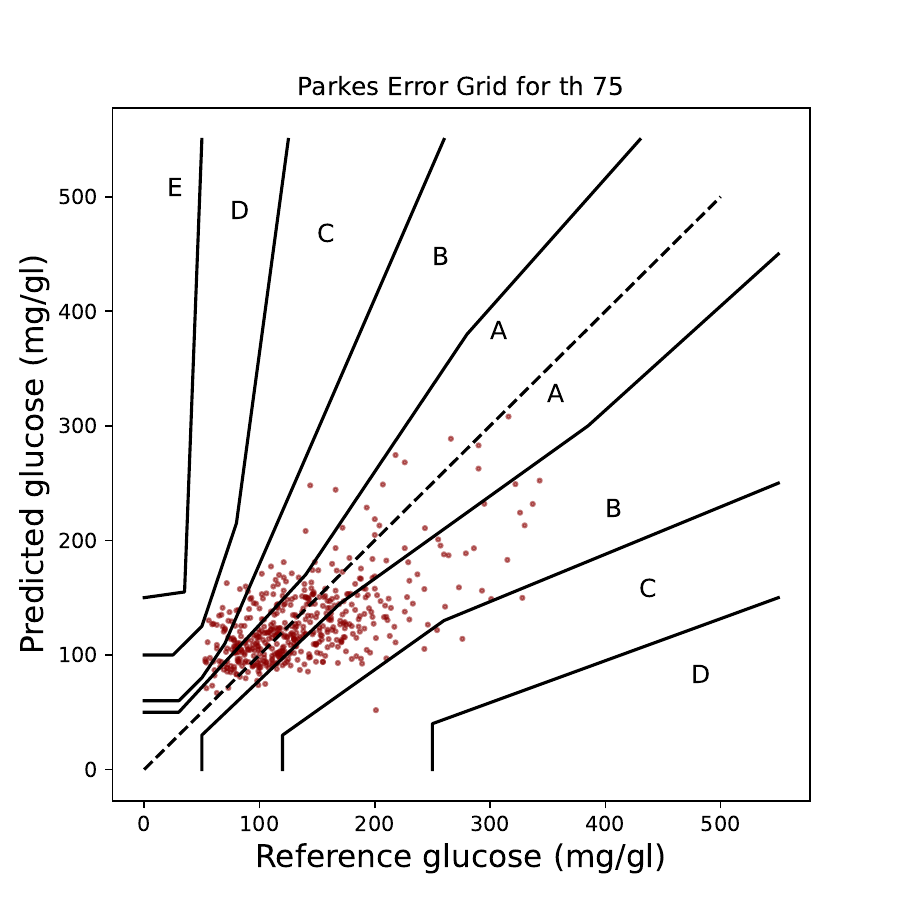}
    \caption{Time horizon 75}
    \label{fig:parkes_th_75_LexGE}
  \end{subfigure}
  \hfill
  \begin{subfigure}{0.49\textwidth}
    \includegraphics[width=\textwidth]{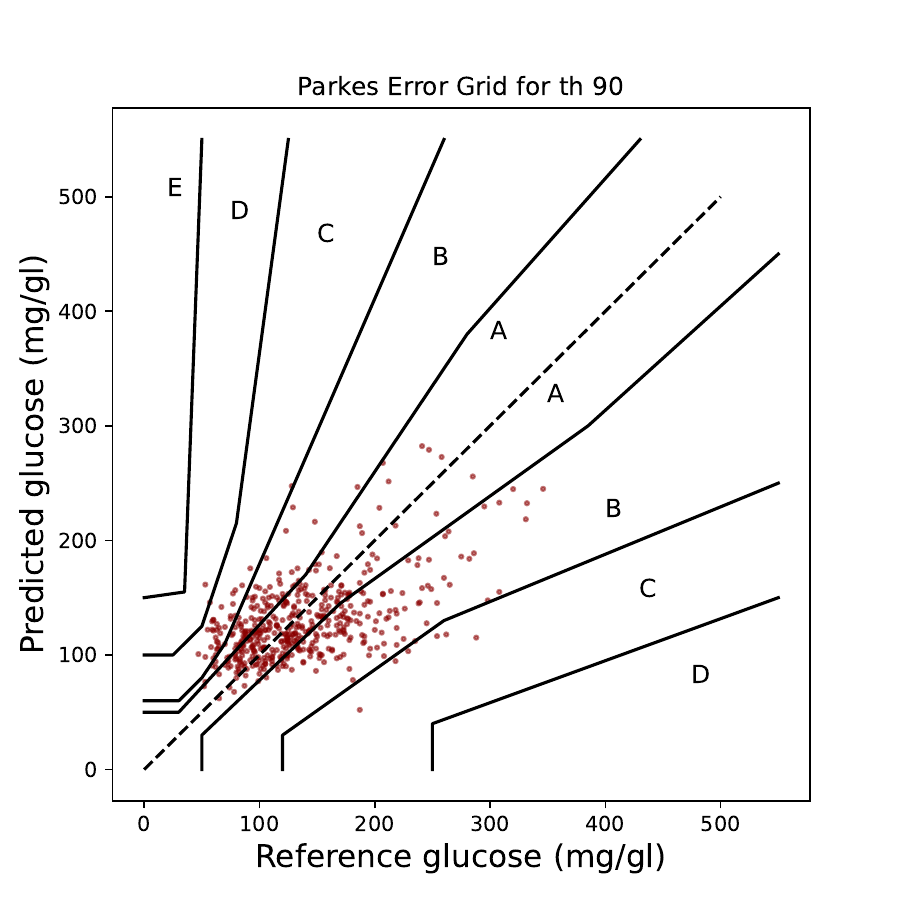}
    \caption{Time horizon 90}
    \label{fig:parkes_th_90_LexGE}
  \end{subfigure}
  \hfill
  \begin{subfigure}{0.49\textwidth}
    \includegraphics[width=\textwidth]{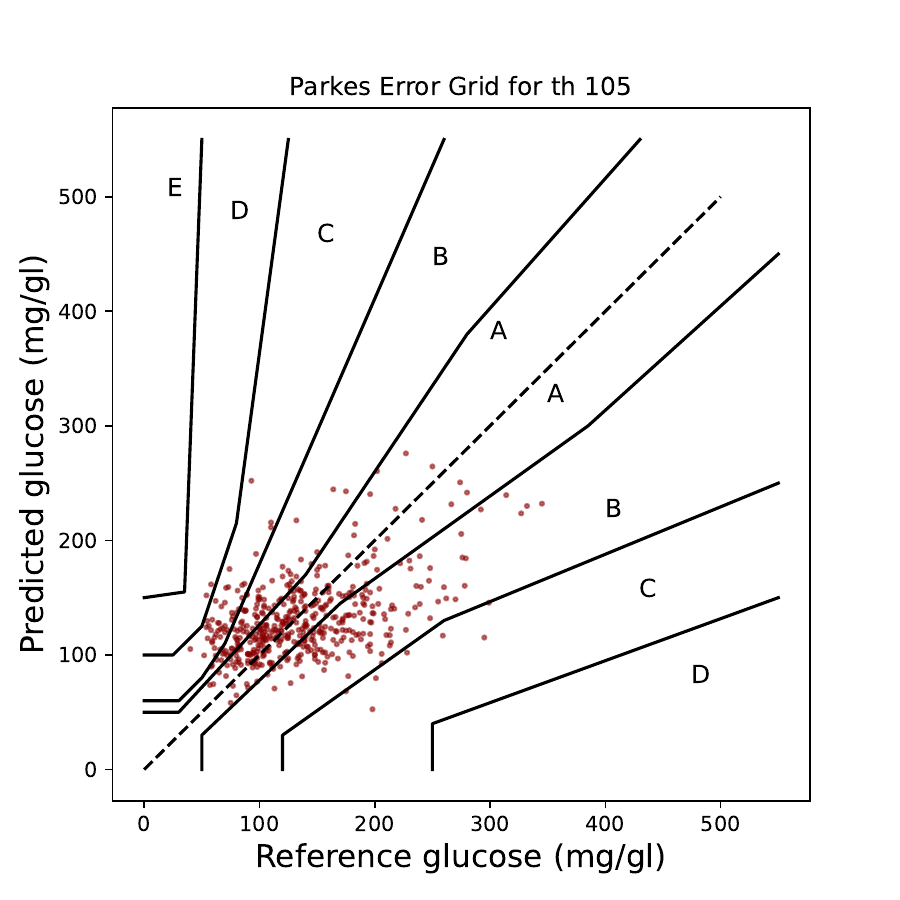}
    \caption{Time horizon 105}
    \label{fig:parkes_th_105_LexGE}
  \end{subfigure}
  \hfill
  \begin{subfigure}{0.49\textwidth}
    \includegraphics[width=\textwidth]{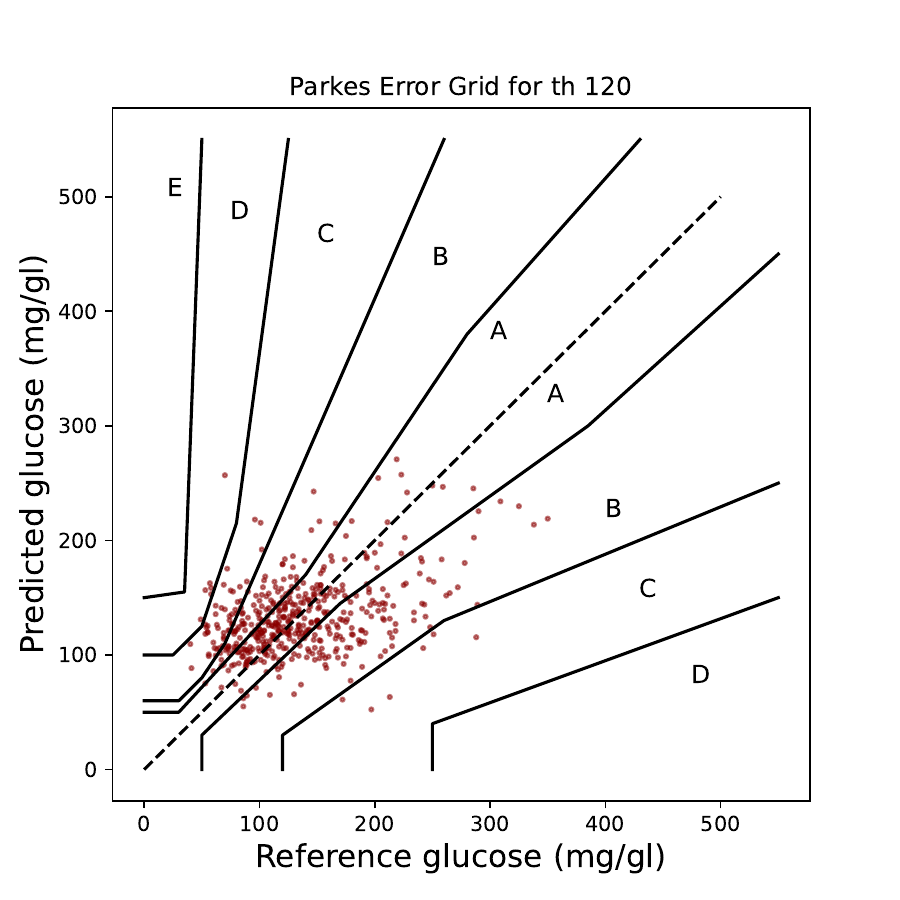}
    \caption{Time horizon 120}
    \label{fig:parkes_th_120_LexGE}
  \end{subfigure}
  \caption{Parkes Errors Grid by Time Horizon for ISIGE, all clusters (part 2).}
  \label{fig:parkes_th_split_LexGE_2}
\end{figure}

Parkes Errors Grid by Time Horizon for $\epsilon$Lexicase ISIGE, all clusters, figures \ref{fig:parkes_th_split_LexGE} and \ref{fig:parkes_th_split_LexGE_2}.

\begin{figure}

\centering
  \begin{subfigure}{0.49\textwidth}
    \includegraphics[width=\textwidth]{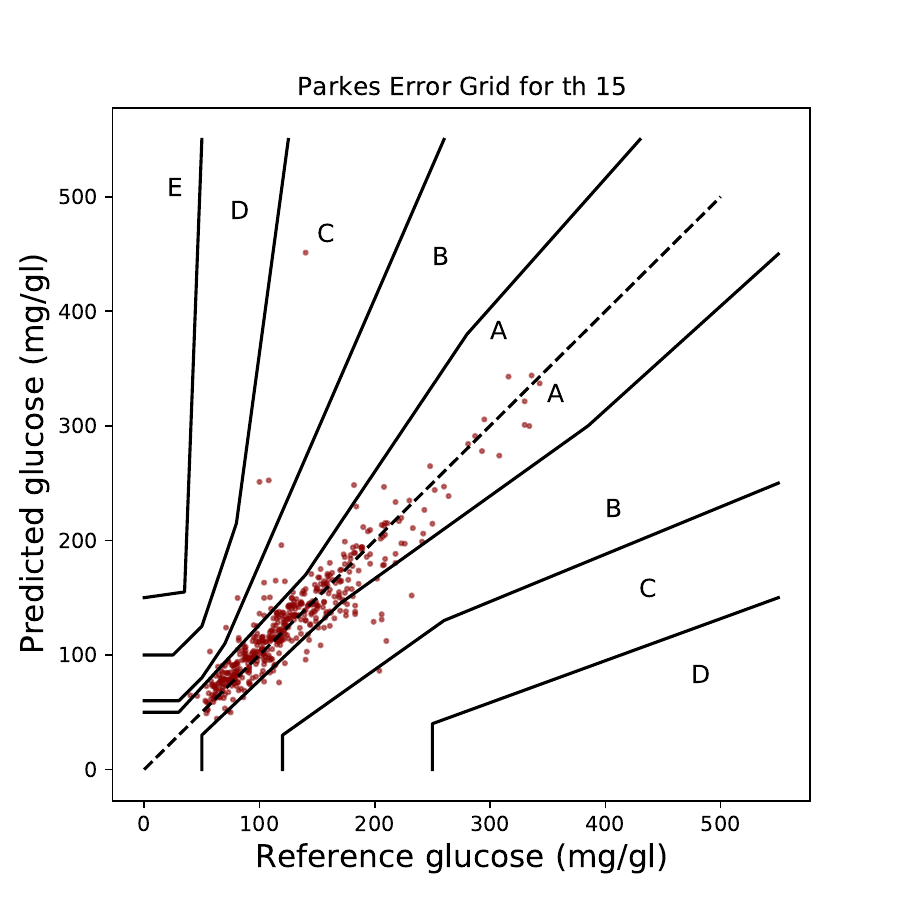}
    \caption{Time horizon 15}
    \label{fig:parkes_th_15_SINDy}
  \end{subfigure}
  \hfill
  \begin{subfigure}{0.49\textwidth}
    \includegraphics[width=\textwidth]{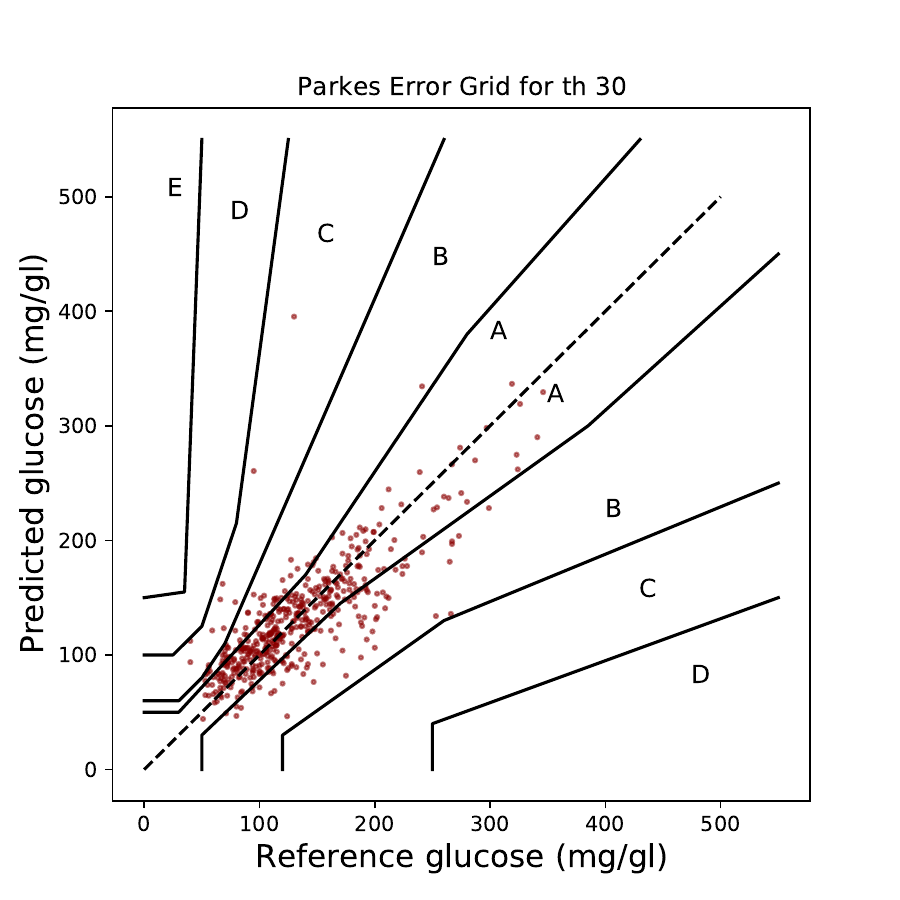}
    \caption{Time horizon 30}
    \label{fig:parkes_th_30_SINDy}
  \end{subfigure}
  \hfill
  \begin{subfigure}{0.49\textwidth}
    \includegraphics[width=\textwidth]{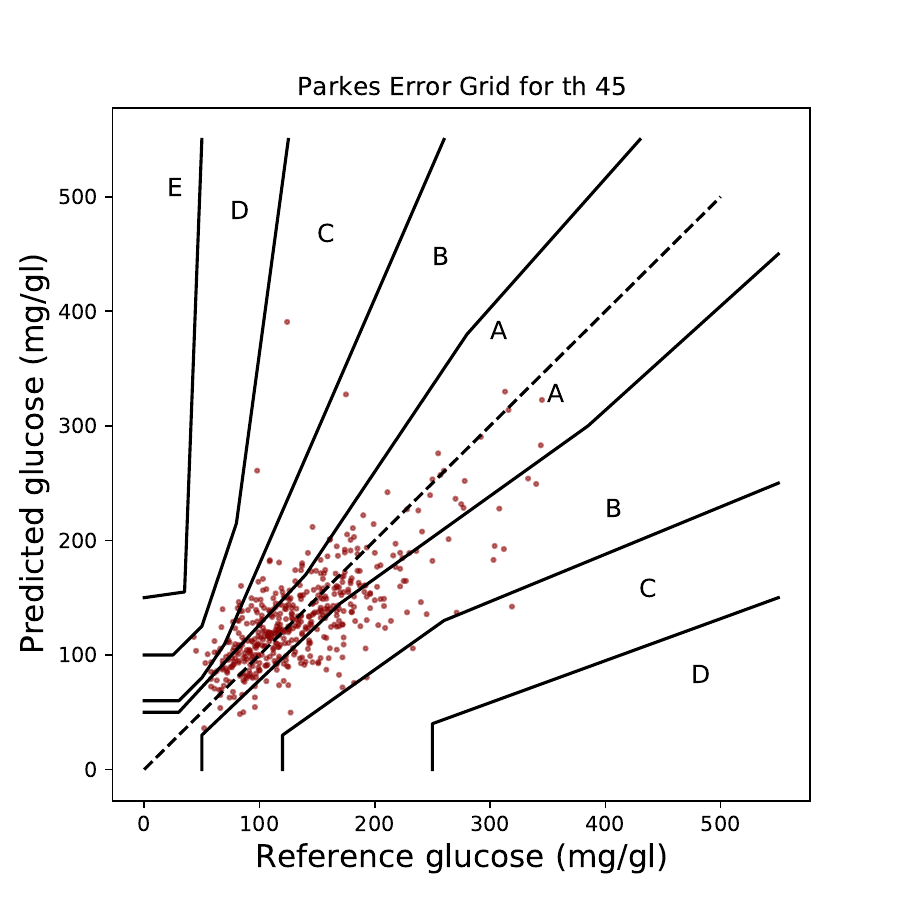}
    \caption{Time horizon 45}
    \label{fig:parkes_th_45_SINDy}
  \end{subfigure}
  \hfill
  \begin{subfigure}{0.49\textwidth}
    \includegraphics[width=\textwidth]{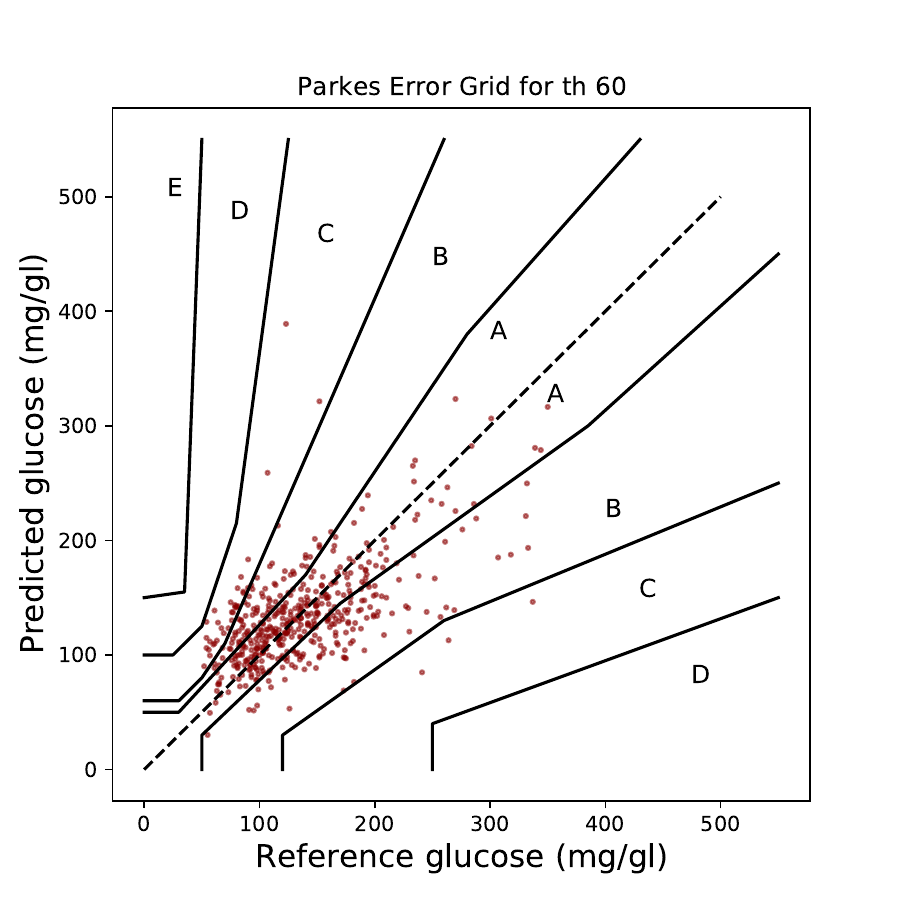}
    \caption{Time horizon 60}
    \label{fig:parkes_th_60_SINDy}
  \end{subfigure}
  \caption{Parkes Errors Grid by Time Horizon for SINDy, all clusters (part 1).}
  \label{fig:parkes_th_split_SINDy}
\end{figure}

\begin{figure}

\centering
  \begin{subfigure}{0.49\textwidth}
    \includegraphics[width=\textwidth]{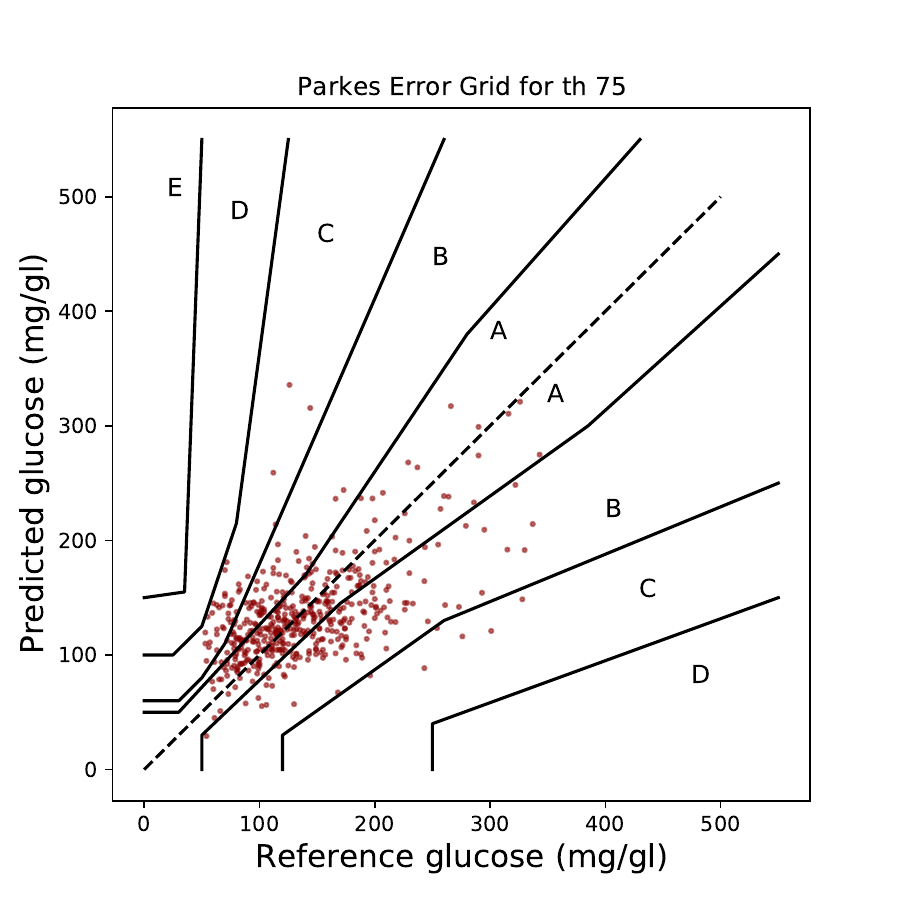}
    \caption{Time horizon 75}
    \label{fig:parkes_th_75_SINDy}
  \end{subfigure}
  \hfill
  \begin{subfigure}{0.49\textwidth}
    \includegraphics[width=\textwidth]{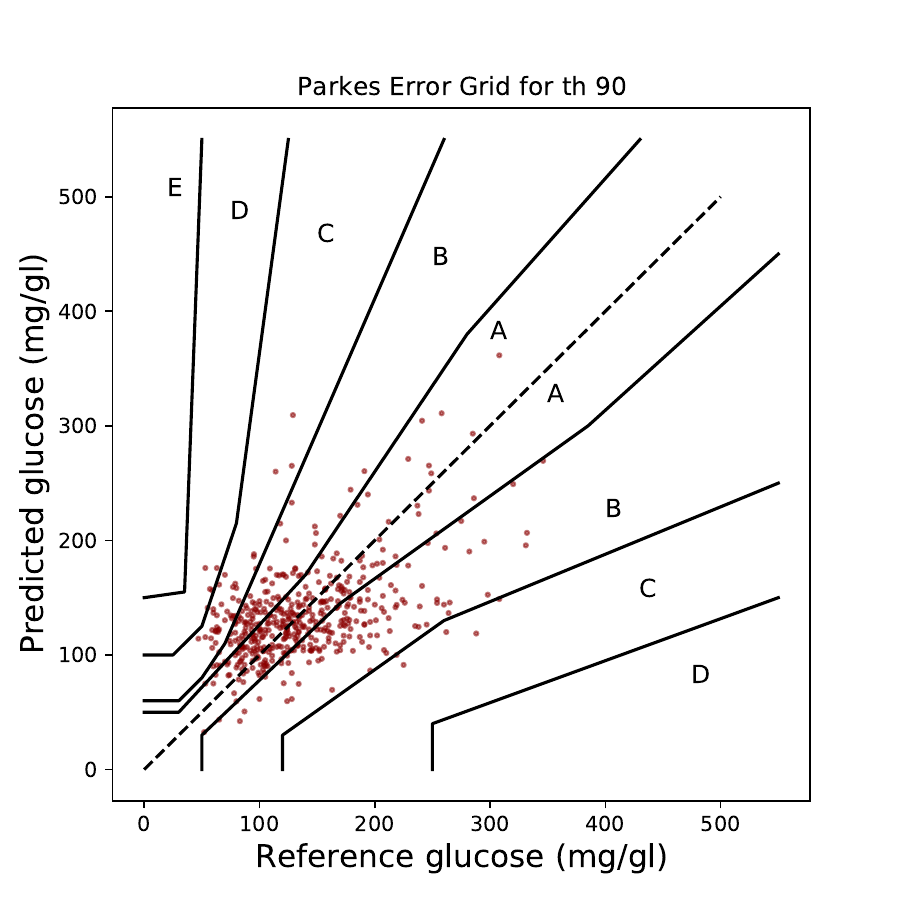}
    \caption{Time horizon 90}
    \label{fig:parkes_th_90_SINDy}
  \end{subfigure}
  \hfill
  \begin{subfigure}{0.49\textwidth}
    \includegraphics[width=\textwidth]{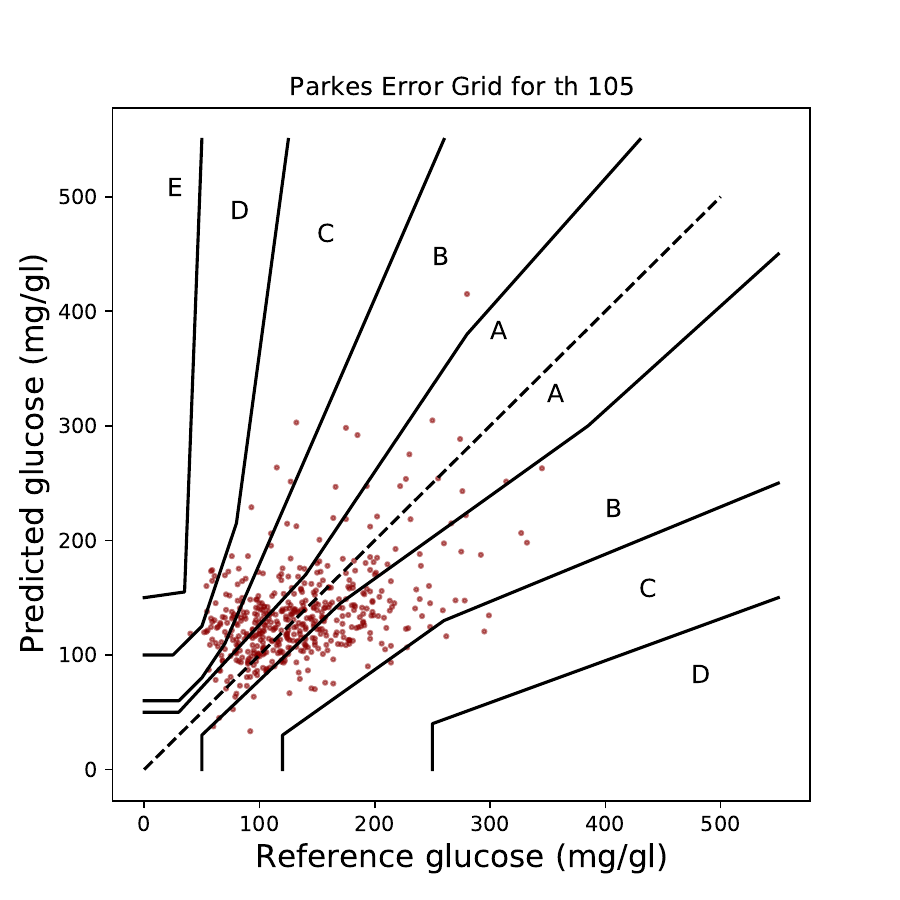}
    \caption{Time horizon 105}
    \label{fig:parkes_th_105_SINDy}
  \end{subfigure}
  \hfill
  \begin{subfigure}{0.49\textwidth}
    \includegraphics[width=\textwidth]{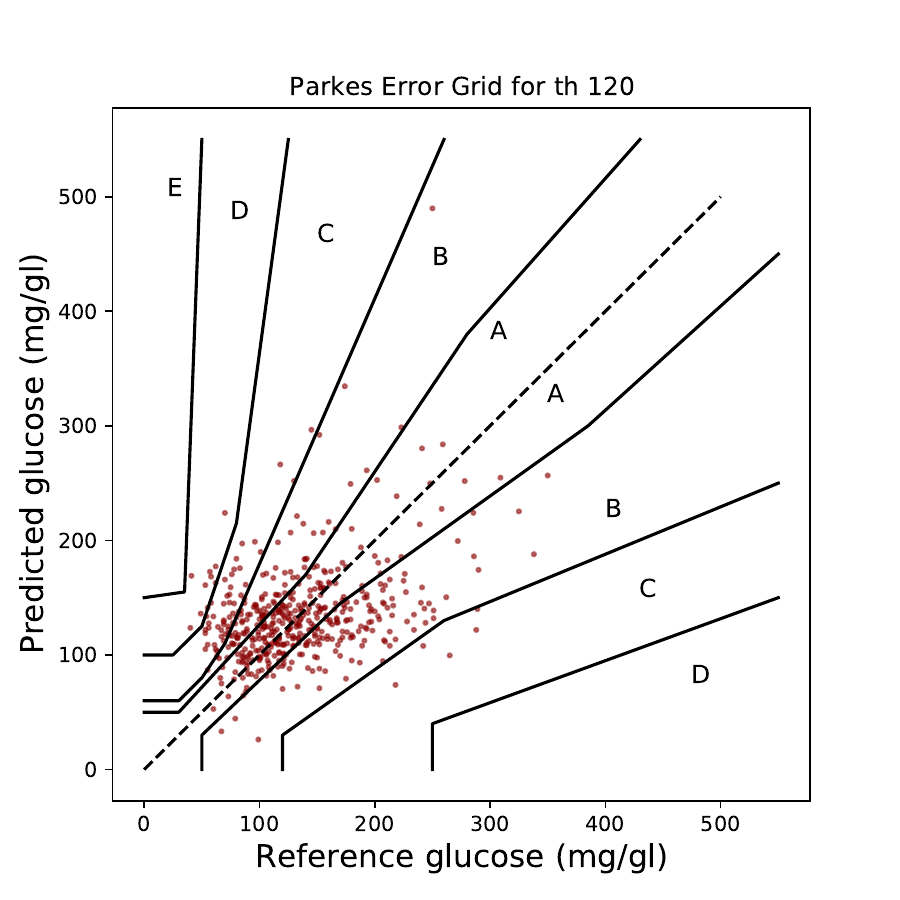}
    \caption{Time horizon 120}
    \label{fig:parkes_th_120_SINDy}
  \end{subfigure}
  \caption{Parkes Errors Grid by Time Horizon for SINDy, all clusters (part 2).}
  \label{fig:parkes_th_split_SINDy_2}
\end{figure}

%-----------------------------------------------
Parkes Errors Grid by Time Horizon for SINDy, all clusters, figures \ref{fig:parkes_th_split_SINDy} and \ref{fig:parkes_th_split_SINDy_2}.

\begin{figure}

\centering
  \begin{subfigure}{0.49\textwidth}
    \includegraphics[width=\textwidth]{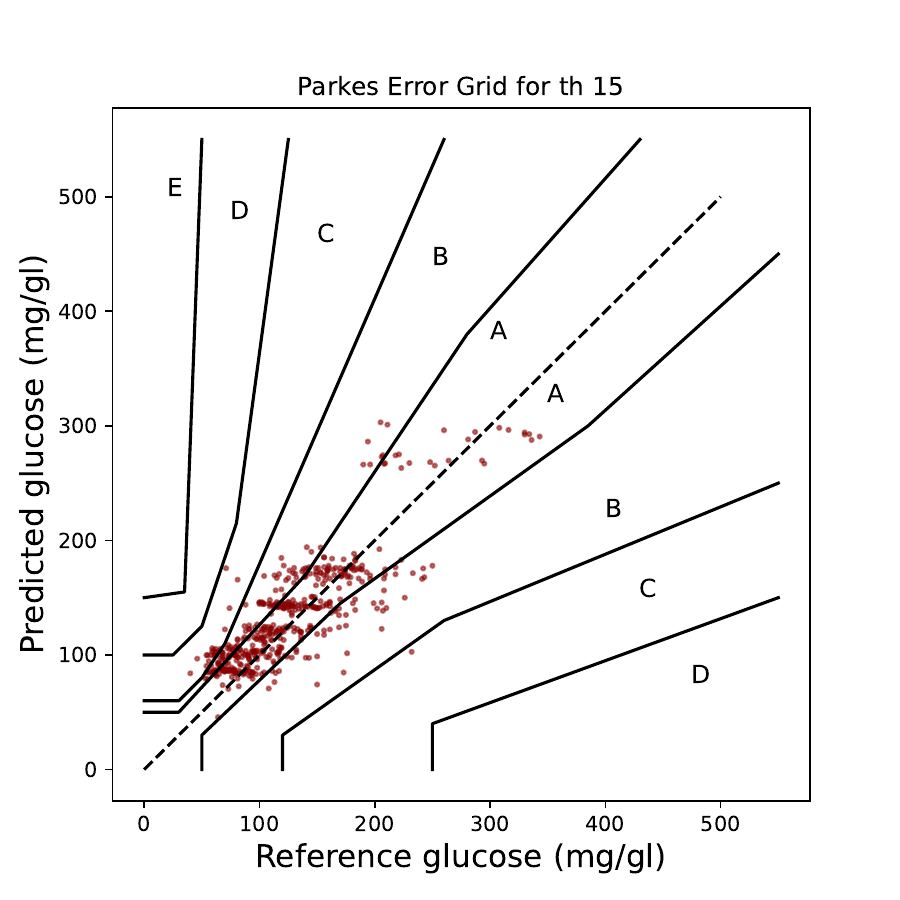}
    \caption{Time horizon 15}
    \label{fig:parkes_th_15_Mirshe}
  \end{subfigure}
  \hfill
  \begin{subfigure}{0.49\textwidth}
    \includegraphics[width=\textwidth]{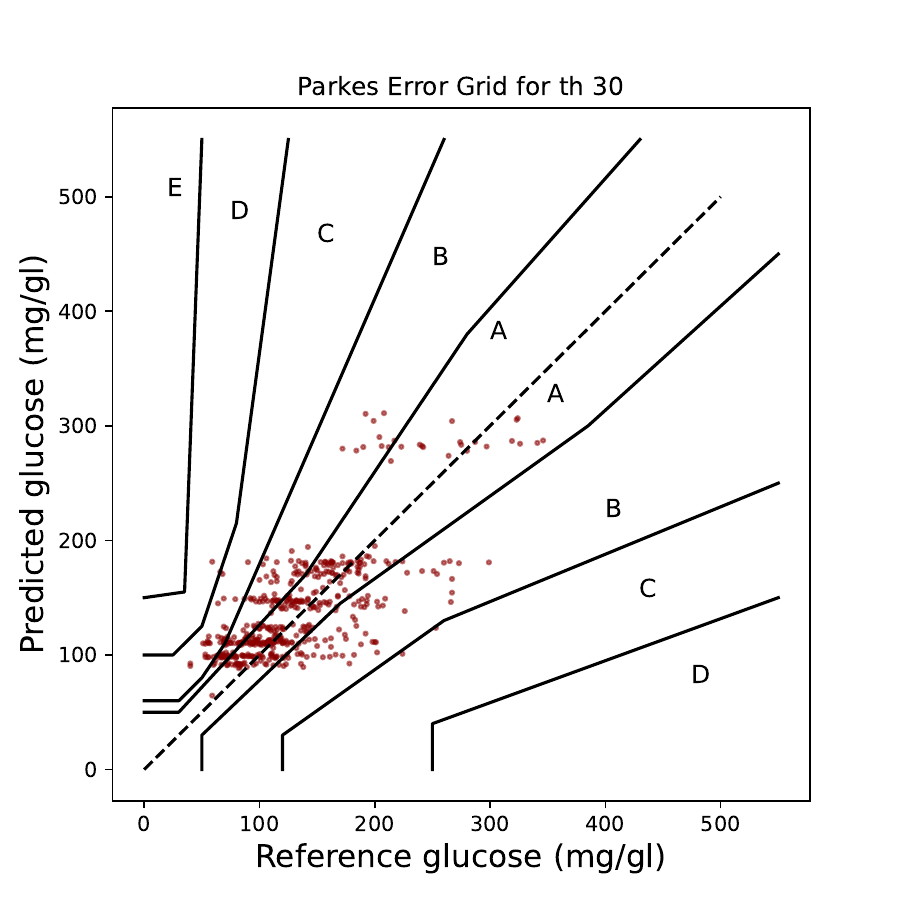}
    \caption{Time horizon 30}
    \label{fig:parkes_th_30_Mirshe}
  \end{subfigure}
  \hfill
  \begin{subfigure}{0.49\textwidth}
    \includegraphics[width=\textwidth]{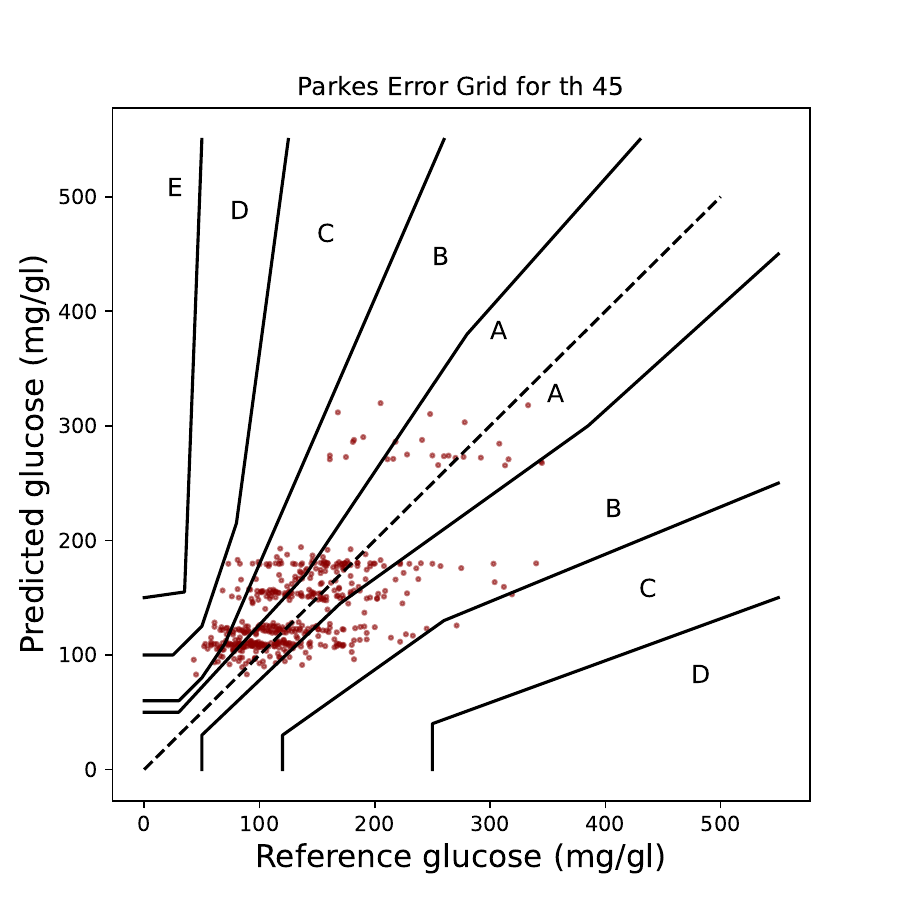}
    \caption{Time horizon 45}
    \label{fig:parkes_th_45_Mirshe}
  \end{subfigure}
  \hfill
  \begin{subfigure}{0.49\textwidth}
    \includegraphics[width=\textwidth]{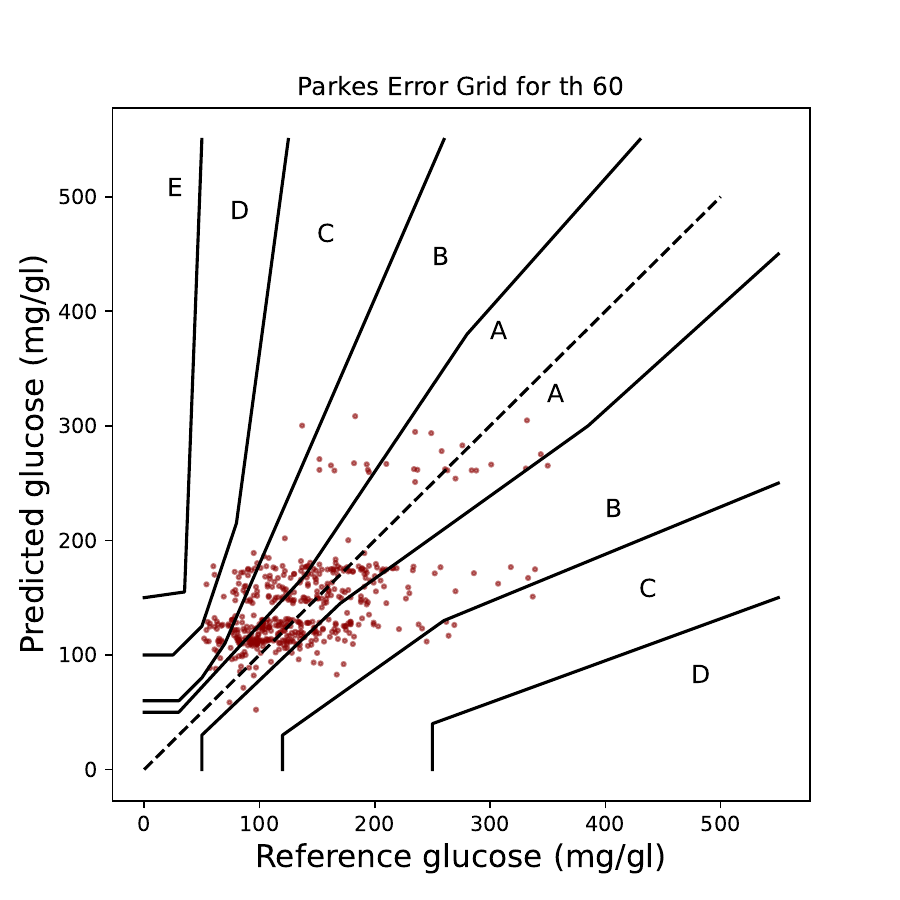}
    \caption{Time horizon 60}
    \label{fig:parkes_th_60_Mirshe}
  \end{subfigure}
  \caption{Parkes Errors Grid by Time Horizon for Mirshekarian, all clusters (part 1).}
  \label{fig:parkes_th_split_Mirshe}
\end{figure}

\begin{figure}

\centering
  \begin{subfigure}{0.49\textwidth}
    \includegraphics[width=\textwidth]{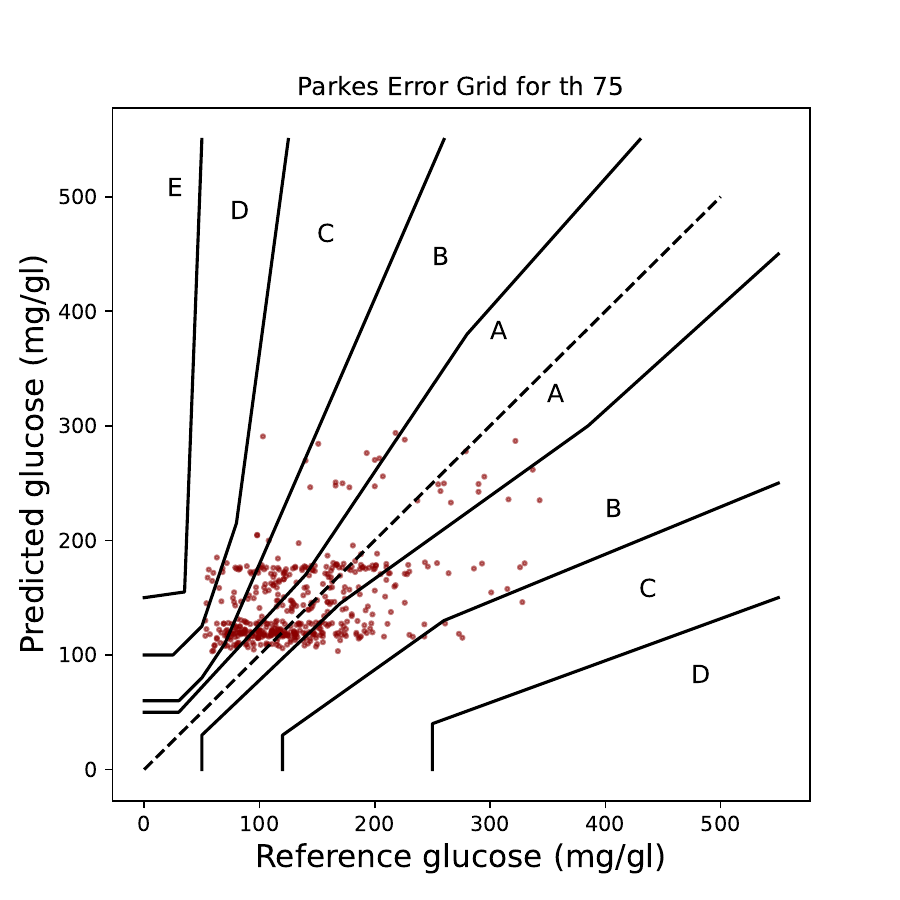}
    \caption{Time horizon 75}
    \label{fig:parkes_th_75_Mirshe}
  \end{subfigure}
  \hfill
  \begin{subfigure}{0.49\textwidth}
    \includegraphics[width=\textwidth]{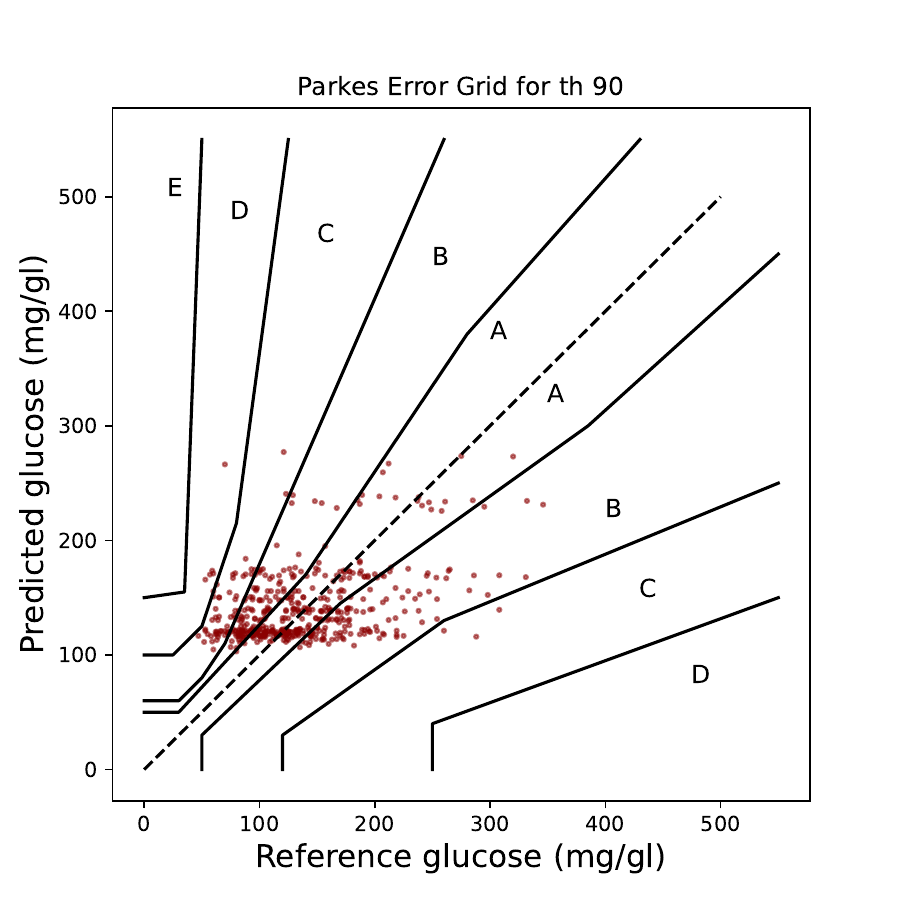}
    \caption{Time horizon 90}
    \label{fig:parkes_th_90_Mirshe}
  \end{subfigure}
  \hfill
  \begin{subfigure}{0.49\textwidth}
    \includegraphics[width=\textwidth]{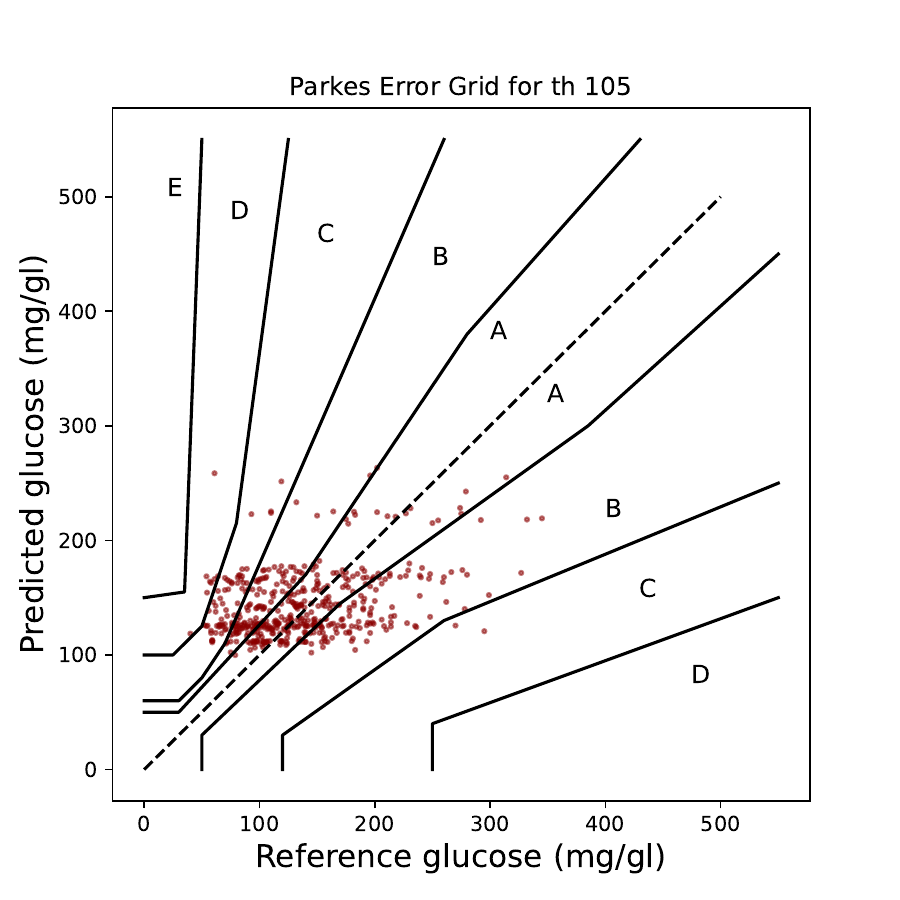}
    \caption{Time horizon 105}
    \label{fig:parkes_th_105_Mirshe}
  \end{subfigure}
  \hfill
  \begin{subfigure}{0.49\textwidth}
    \includegraphics[width=\textwidth]{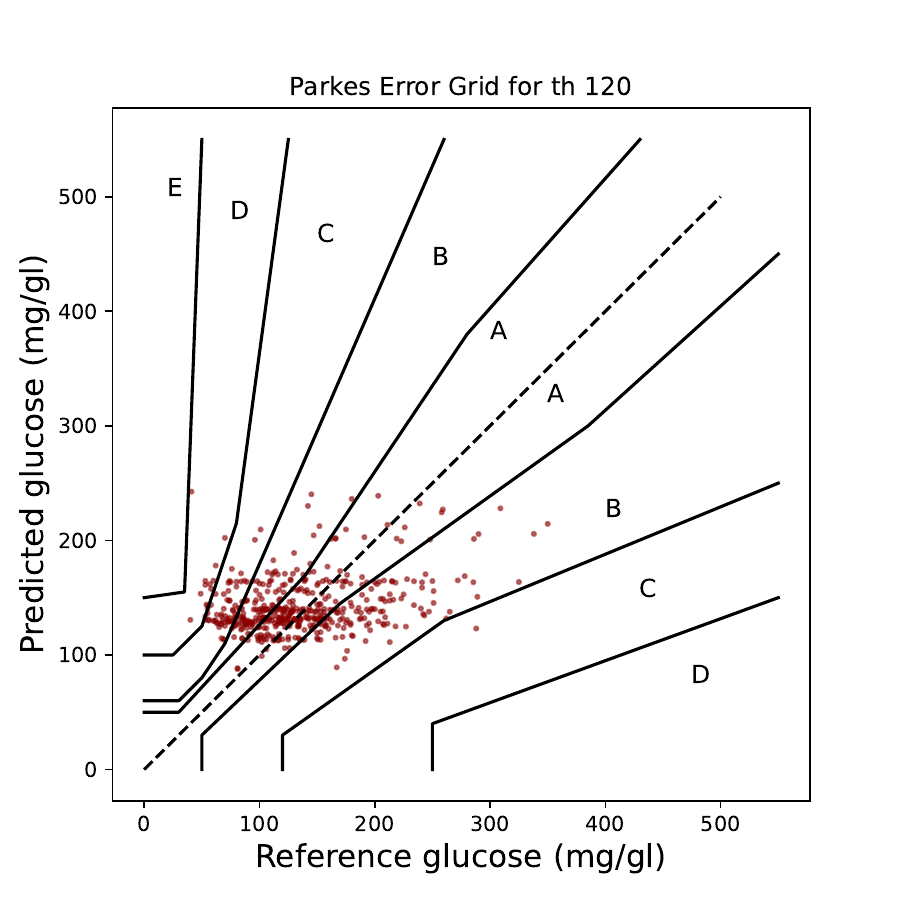}
    \caption{Time horizon 120}
    \label{fig:parkes_th_120_Mirshe}
  \end{subfigure}
  \caption{Parkes Errors Grid by Time Horizon for Mirshekarian, all clusters (part 2).}
  \label{fig:parkes_th_split_Mirshe_2}
\end{figure}

%-----------------------------------------------
Parkes Errors Grid by Time Horizon for SINDy, all clusters, figures \ref{fig:parkes_th_split_Mirshe} and \ref{fig:parkes_th_split_Mirshe_2}.
\newpage
\newpage
\section{Expressions per cluster}
\label{aped.B}

Table \ref{tab:Expression_SINDy} shows the expressions obtained by SINDy for each cluster.
\begin{table}[h]
\resizebox{\textwidth}{!}{
\begin{tabular}{cl}
Cluster & \multicolumn{1}{c}{Expression} \\
1 & $ G'(t) = 181.854 - 2706.076 \cdot I_{Bg}(t) + 3814.515 \cdot C_{Bg}(t) - 637.866 \cdot B_{I}(t) - 152.312 \cdot T_{M}(t) \cdot I_{Bg}(t) + 19.558 \cdot T_{M}(t) \cdot C_{Bg}(t) $ \\ 
 & $ + 10.907 \cdot T_{M}(t) \cdot B_{I}(t) + 37.952 \cdot I_{Bg}(t) \cdot c_{60}(t) + 25.308 \cdot I_{Bg}(t) \cdot H_{60}(t) - 3423.206 \cdot C_{Bg}(t)^2 $ \\ 
 & $ - 73.652 \cdot C_{Bg}(t) \cdot c_{60}(t) - 26.789 \cdot C_{Bg}(t) \cdot H_{60}(t) + 2.087 \cdot C_{Bg}(t) \cdot S_{60}(t) + 2319.862 \cdot C_{Bg}(t) \cdot B_{I}(t) $ \\ 
 & $ + 2010.550 \cdot B_{I}(t)^2 $ \\ 
2 & $ G'(t) =  - 23.920 + 1.026 \cdot G(t-1) + 4.753 \cdot T_{M}(t) + 4.167 \cdot I_{Bm}(t) + 59.537 \cdot B_{I}(t) - 14.457 \cdot T_{M}(t) \cdot B_{I}(t) $ \\ 
 & $ - 9.834 \cdot I_{Bm}(t) \cdot B_{I}(t) - 14.297 \cdot B_{I}(t)^2 $ \\ 
3 & $ G'(t) = 0.980 \cdot G(t-1) + 0.135 \cdot T_{M}(t) \cdot G''(t-1) $ \\ 
4 & $ G'(t) = 1.042 \cdot G(t-1) - 0.106 \cdot H_{30}(t) $ \\ 
5 & $ G'(t) = 9.332 + 1.014 \cdot G(t-1) - 2.449 \cdot T_{M}(t) - 7.165 \cdot B_{I}(t) + 10.019 \cdot T_{M}(t) \cdot B_{I}(t) + 2.430 \cdot I_{Bm}(t) \cdot B_{I}(t) $ \\ 
 & $ - 70.013 \cdot B_{I}(t)^2 $ \\ 
6 & $ G'(t) = 24.734 + 0.905 \cdot G(t-1) - 1.596 \cdot T_{M}(t) + 1.219 \cdot I_{Bm}(t) - 47.738 \cdot B_{I}(t) + 0.362 \cdot G(t-1) \cdot B_{I}(t) $ \\ 
 & $ - 0.497 \cdot T_{M}(t) \cdot I_{Bm}(t) + 0.115 \cdot T_{M}(t) \cdot C_{BM}(t) + 1.851 \cdot T_{M}(t) \cdot B_{I}(t) - 0.231 \cdot I_{Bm}(t)^2 $ \\ 
 & $ + 9.336 \cdot I_{Bm}(t) \cdot B_{I}(t) - 0.662 \cdot C_{BM}(t) \cdot B_{I}(t) + 0.476 \cdot c_{30}(t) \cdot B_{I}(t) - 0.540 \cdot H_{30}(t) \cdot B_{I}(t) $ \\ 
 & $ - 16.124 \cdot B_{I}(t)^2 $ \\ 
7 & $ G'(t) = 0.995 \cdot G(t-1) - 0.243 \cdot G(t-1) \cdot I_{Bg}(t) + 0.130 \cdot T_{M}(t) \cdot G'(t-1) $ \\ 
8 & $ G'(t) = 1.023 \cdot G(t-1) - 4.029 \cdot G(t-1) \cdot C_{Bg}(t-30) + 1.978 \cdot G(t-1) \cdot I_{Bg}(t-45) - 1.251 \cdot G(t-1) \cdot C_{Bg}(t-45) + 7.035 \cdot H_{30}(t) \cdot C_{Bg}(t-30) $ \\ 
 & $ - 3.142 \cdot H_{30}(t) \cdot I_{Bg}(t-45) $ \\ 
9 & $ G'(t) = 1.032 \cdot G(t-1) - 2.572 \cdot G(t-1) \cdot C_{Bg}(t) $ \\ 
10 & $ G'(t) = 1.050 \cdot G(t-1) + 4.970 \cdot G(t-1) \cdot I_{Bg}(t-15) - 5.794 \cdot G(t-1) \cdot I_{Bg}(t-30) - 1.373 \cdot G(t-1) \cdot I_{Bg}(t-45) $ \\ 
 & $ - 7.067 \cdot H_{30}(t) \cdot I_{Bg}(t-15) + 2.505 \cdot H_{30}(t) \cdot C_{Bg}(t-15) + 7.262 \cdot H_{30}(t) \cdot I_{Bg}(t-30) $ \\ 
11 & $ G'(t) = 1.556 \cdot G(t-1) - 0.916 \cdot H_{30}(t) - 1.497 \cdot G'(t-1) - 0.218 \cdot G''(t-1) - 0.165 \cdot G(t-1) \cdot T_{M}(t) - 10.700 \cdot G(t-1) \cdot I_{Bg}(t) $ \\ 
 & $ + 4.396 \cdot G(t-1) \cdot C_{Bg}(t) + 0.377 \cdot G(t-1) \cdot B_{I}(t) + 0.745 \cdot T_{M}(t)^2 + 0.204 \cdot T_{M}(t) \cdot H_{30}(t) + 0.198 \cdot T_{M}(t) \cdot G''(t-1) $ \\ 
 & $ - 9.179 \cdot I_{Bg}(t) \cdot c_{30}(t) + 16.677 \cdot I_{Bg}(t) \cdot H_{30}(t) + 2.197 \cdot I_{Bg}(t) \cdot S_{30}(t) + 9.326 \cdot I_{Bg}(t) \cdot G'(t-1) $ \\ 
 & $ + 4.338 \cdot I_{Bg}(t) \cdot G''(t-1) - 4.349 \cdot C_{Bg}(t) \cdot H_{30}(t) - 1.192 \cdot C_{Bg}(t) \cdot S_{30}(t) - 0.183 \cdot S_{30}(t) \cdot B_{I}(t) $ \\ 
12 & $ G'(t) = 1.349 \cdot G(t-1) - 0.644 \cdot H_{30}(t) - 4.321 \cdot G(t-1) \cdot C_{Bg}(t) - 5.870 \cdot G(t-1) \cdot I_{Bg}(t-15) + 1.481 \cdot G(t-1) \cdot C_{Bg}(t-15) $ \\ 
 & $ + 3.010 \cdot G(t-1) \cdot I_{Bg}(t-30) - 2.549 \cdot G(t-1) \cdot C_{Bg}(t-30) - 0.995 \cdot G(t-1) \cdot I_{Bg}(t-45) + 0.152 \cdot G(t-1) \cdot I_{Bg}(t-60) $ \\ 
 & $ - 6.401 \cdot G(t-1) \cdot C_{Bg}(t-60) + 5.004 \cdot C_{Bg}(t) \cdot H_{30}(t) + 8.597 \cdot H_{30}(t) \cdot I_{Bg}(t-15) + 1.356 \cdot H_{30}(t) \cdot C_{Bg}(t-15) $ \\ 
 & $ - 1.164 \cdot H_{30}(t) \cdot I_{Bg}(t-45) + 12.415 \cdot H_{30}(t) \cdot C_{Bg}(t-60) $ \\ 
13 & $ G'(t) = 6.897 + 1.041 \cdot G(t-1) - 0.743 \cdot T_{M}(t) - 0.270 \cdot C_{BM}(t) - 0.276 \cdot G(t-1) \cdot B_{I}(t) + 0.235 \cdot T_{M}(t) \cdot I_{Bm}(t) $ \\ 
 & $ - 1.196 \cdot I_{Bm}(t) \cdot B_{I}(t) + 0.845 \cdot C_{BM}(t) \cdot B_{I}(t) + 25.383 \cdot B_{I}(t)^2 $ \\ 
14 & $ G'(t) = 43.211 + 0.968 \cdot G(t-1) - 1.323 \cdot T_{M}(t) + 8.345 \cdot I_{Bm}(t) - 0.442 \cdot C_{BM}(t) + 0.191 \cdot c_{30}(t) - 0.613 \cdot H_{30}(t) $ \\ 
 & $ - 29.836 \cdot B_{I}(t) + 0.280 \cdot T_{M}(t) \cdot B_{I}(t) - 47.855 \cdot I_{Bm}(t) \cdot B_{I}(t) + 3.367 \cdot C_{BM}(t) \cdot B_{I}(t) - 0.426 \cdot c_{30}(t) \cdot B_{I}(t) $ \\ 
 & $ + 1.669 \cdot H_{30}(t) \cdot B_{I}(t) - 181.578 \cdot B_{I}(t)^2 $ \\ 
15 & $ G'(t) = 2.064 + 0.906 \cdot G(t-1) - 6.279 \cdot T_{M}(t) - 404.138I_{Bm}(t) + 59.925 \cdot C_{Bg}(t) - 0.257 \cdot c_{30}(t) + 0.446 \cdot h_{30}(t) $ \\ 
 & $ + 205.460 \cdot B_{I}(t) + 1.041 \cdot G'(t-1) + 3.971 G(t-1) \cdot I_{Bm}(t) - 0.275 \cdot G(t-1) \cdot C_{Bg}(t) - 0.650 \cdot G(t-1) \cdot B_{I}(t) $ \\
 & $ + 0.527 \cdot T_{M}(t)^2 + 67.439 \cdot T_{M}(t) \cdot I_{Bm}(t) - 27.930 \cdot T_{M}(t) \cdot C_{Bg}(t) + 0.862 \cdot T_{M}(t) \cdot B_{I}(t) + 2600.891 \cdot I_{Bm}(t)^2 $ \\
 & $ + 3515.204 \cdot I_{Bm}(t) \cdot C_{Bg}(t) + 9.307I_{Bm}(t) \cdot c_{30}(t) - 10.954 \cdot I_{Bm}(t) \cdot h_{30}(t) - 0.281I_{Bm}(t) \cdot S_{30}(t) $ \\
 & $ - 7.904I_{Bm}(t) \cdot B_{I}(t) + 4.365I_{Bm}(t) \cdot G'(t-1) - 6.653 \cdot I_{Bm}(t) \cdot G''(t-1) - 1085.326 \cdot C_{Bg}(t)^2 - 8.024 \cdot C_{Bg}(t) \cdot c_{30}(t) $ \\
 & $ + 3.287 \cdot C_{Bg}(t) \cdot h_{30}(t) + 0.523 \cdot C_{Bg}(t) \cdot S_{30}(t) - 15.630 \cdot C_{Bg}(t) \cdot G'(t-1) + 7.066 \cdot C_{Bg}(t) \cdot G''(t-1) $ \\
 & $ + 1.109 \cdot c_{30}(t) \cdot B_{I}(t) - 1.826 \cdot h_{30}(t) \cdot B_{I}(t) - 128.624 \cdot B_{I}(t)^2 - 1.342 \cdot B_{I}(t) \cdot G'(t-1) $

\end{tabular}
} 
\caption{Expressions for each cluster using SINDy}
\label{tab:Expression_SINDy}
\end{table}

\begin{comment}

\begin{figure}
    \centering
\begin{lstlisting}[mathescape]

<func> ::= G(t_n) + <expr2>

<expr2> ::= (<expr2> <op> <expr2>) | (<cte> <op> <var2>  <op> <expr2>)| <var2> | (-<var2>) | pow(<var2>,<sign><exponent>)|
(-pow(<var2>,<sign><exponent>))

<var2> ::= $B_I(t_n)| I_B(t_n)|F_{ch}(t_n)|\text{HR}(t_n)|C(t_n)|S(t_n)|$
$G(t_n)*G(t_n)|G(t_n)*B_I(t_n)|G(t_n)*I_B(t_n)|G(t_n)*F_{ch}(t_n)|G(t_n)*\text{HR}(t_n)|G(t_n)*C(t_n)|G(t_n)*S(t_n)|$
$B_I(t_n)*B_I(t_n)|B_I(t_n)*I_B(t_n)|B_I(t_n)*F_{ch}(t_n)|B_I(t_n)*\text{HR}(t_n)|B_I(t_n)*C(t_n)|B_I(t_n)*S(t_n)|$
$I_B(t_n)*I_B(t_n)|I_B(t_n)*F_{ch}(t_n)|I_B(t_n)*\text{HR}(t_n)|I_B(t_n)*C(t_n)|I_B(t_n)*S(t_n)|$
$F_{ch}(t_n)*F_{ch}(t_n)|F_{ch}(t_n)*\text{HR}(t_n)|F_{ch}(t_n)*C(t_n)|F_{ch}(t_n)*S(t_n)|$
$\text{HR}(t_n)*\text{HR}(t_n)|\text{HR}(t_n)*C(t_n)|\text{HR}(t_n)*S(t_n)|$
$C(t_n)*C(t_n)|C(t_n)*S(t_n)|$
$S(t_n)*S(t_n)$

<op> ::= +|-|*

<cte> ::= <base>*pow(10,<sign><exponent>)

<base> ::= 1|2|3|4| $\cdot \cdot \cdot$ |99

<exponent> ::= 1|2|3|4|5|6|8|9

<sign> ::= +|-

\end{lstlisting}  \caption{Grammars used for $\epsilon$-Lexicase ISIGE.}
    \label{fig:grammar}
\end{figure}
\end{comment}
\newpage

%%%%%%%%%%%%%%%%%%%%%%%%%%%%%%%%%%%%%%%%%%%%%%%%
%%%%%%%%%%%%%%%%%%%%%%%%%%%%%%%%%%%%%%%%%%%%%%%%
\newpage

\end{document}